%% file: main.tex
\documentclass[journal,transmag]{IEEEtran}
\IEEEoverridecommandlockouts

\usepackage{graphicx} 
\usepackage{amsmath}
\usepackage{amssymb}
\usepackage{caption}
\usepackage{amsfonts}
\usepackage{textcomp}
\usepackage{xcolor}
\usepackage{soul}
\usepackage{multirow}
\usepackage{array}
\usepackage{hhline}
\usepackage{titlesec}
\usepackage{stfloats}
\usepackage{url}
\usepackage{verbatim}
\usepackage{tikz}
\mathchardef\mhyphen="2D
\usepackage{color}
\usepackage{microtype}
\usepackage{wrapfig}
\usepackage{booktabs} 
\usepackage{dsfont}
\usepackage{pgfplots}
\usetikzlibrary{patterns}
\usepgfplotslibrary{fillbetween}
\usepackage{pgfkeys}
\usepackage{lipsum}
\usepackage{mwe}
\usepackage{tikz}
\usetikzlibrary{backgrounds}
\usetikzlibrary{intersections}
\usetikzlibrary{shapes.geometric}
\usetikzlibrary{spy}
\usetikzlibrary{shapes.geometric}
\usepackage{pgfplotstable}
\pgfmathdeclarefunction{gauss}{2}{%
  \pgfmathparse{1/(#2*sqrt(2*pi))*exp(-((x-#1)^2)/(2*#2^2))}%
}
\pgfplotsset{compat=1.11,
    /pgfplots/ybar legend/.style={
    /pgfplots/legend image code/.code={%
       \draw[##1,/tikz/.cd,yshift=-0.25em]
        (0cm,0cm) rectangle (3pt,0.8em);},
   },
}
\usepackage{boldline}
\usepackage{colortbl}
\usepgfplotslibrary{colormaps}
    \def\addlegendimage{\csname pgfplots@addlegendimage\endcsname}

\pgfplotsset{ 
cycle list={%
{draw=black,mark=star,solid},
{draw=black, mark=square,solid}}}

\usepackage{algorithm}
\usepackage{algorithmicx}
\usepackage{algpseudocode}
\algrenewcommand\algorithmicrequire{\textbf{Input:}}
\algrenewcommand\algorithmicensure{\textbf{Output:}}
\newcommand{\brokenline}[2][t]{\parbox[#1]{\dimexpr\linewidth-\ALG@thistlm}{\strut\raggedright #2\strut}}

\usepackage{hyperref}
\hypersetup{
    colorlinks=true,
    linkcolor=blue,
    filecolor=magenta,      
    urlcolor=cyan,
    pdftitle={Overleaf Example},
    pdfpagemode=FullScreen,
    }
\usepackage{subcaption}

\begin{document}

%\title{DeepVigor+: A Scalable and Accurate Semi-Analytical Method for Fault Resilience Analysis of Deep Neural Networks}
\title{DeepVigor+: Scalable and Accurate Semi-Analytical Fault Resilience Analysis for Convolutional Neural Networks \vspace{-2mm}}

\author{
\IEEEauthorblockN{
Mohammad Hasan Ahmadilivani\IEEEauthorrefmark{1},
Jaan Raik\IEEEauthorrefmark{1},
Masoud Daneshtalab\IEEEauthorrefmark{1,2}, 
Maksim Jenihhin\IEEEauthorrefmark{1}}
\IEEEauthorblockA{\IEEEauthorrefmark{1}Tallinn University of Technology, Tallinn, Estonia}
\IEEEauthorblockA{\IEEEauthorrefmark{2}Mälardalen University, Västerås, Sweden \vspace{-4mm}}
}

\markboth{Submitted to an IEEE Transactions}%
{M. H. Ahmadilivani \MakeLowercase{\textit{et al.}}: DeepVigor+: Scalable and Accurate Semi-Analytical Fault Resilience Analysis for Deep Neural Networks}

\IEEEtitleabstractindextext{%
\begin{abstract}

The growing exploitation of Machine Learning (ML) in safety-critical applications necessitates rigorous safety analysis. Hardware reliability assessment is a major concern with respect to measuring the level of safety in ML-based systems. Quantifying the reliability of emerging ML models, including Convolutional Neural Networks (CNNs), is highly complex due to their enormous size in terms of the number of parameters and computations. Conventionally, Fault Injection (FI) is applied to perform a reliability measurement. However, performing FI on modern-day CNNs is prohibitively time-consuming if an acceptable confidence level is to be achieved. To speed up FI for large CNNs, statistical FI (SFI) has been proposed, but its runtimes are still considerably long.

In this work, we introduce DeepVigor+, a scalable, fast, and accurate semi-analytical method as an efficient alternative for reliability measurement in CNNs. DeepVigor+ implements a fault propagation analysis model and attempts to acquire Vulnerability Factors (VFs) as reliability metrics in an optimal way. The results indicate that DeepVigor+ obtains VFs for CNN models with an error less than $1\%$, i.e., the objective in SFI, but with $14.9$ up to $26.9$ times fewer simulations than the best-known state-of-the-art SFI. DeepVigor+ enables an accurate reliability analysis for large and deep CNNs within a few minutes, rather than achieving the same results in days or weeks.  \vspace{-2em}

\end{abstract}

% \begin{IEEEkeywords}
% Reliability, Resilience Analysis, Deep Neural Network
% \end{IEEEkeywords}
}

\maketitle

\IEEEdisplaynontitleabstractindextext

\input{sections/1-introduction}
\input{sections/2-related-works}
\input{sections/3-method}

\input{sections/4-setup}

\input{sections/5-results}
\input{sections/6-conclusion}

%\section*{}
\bibliographystyle{IEEEtran}
\bibliography{refs.bib}
\vspace{-10mm}

%\newpage
%\vfill
\newpage

\begin{IEEEbiography}[{\includegraphics[width=1in,height=1.25in,clip,keepaspectratio]{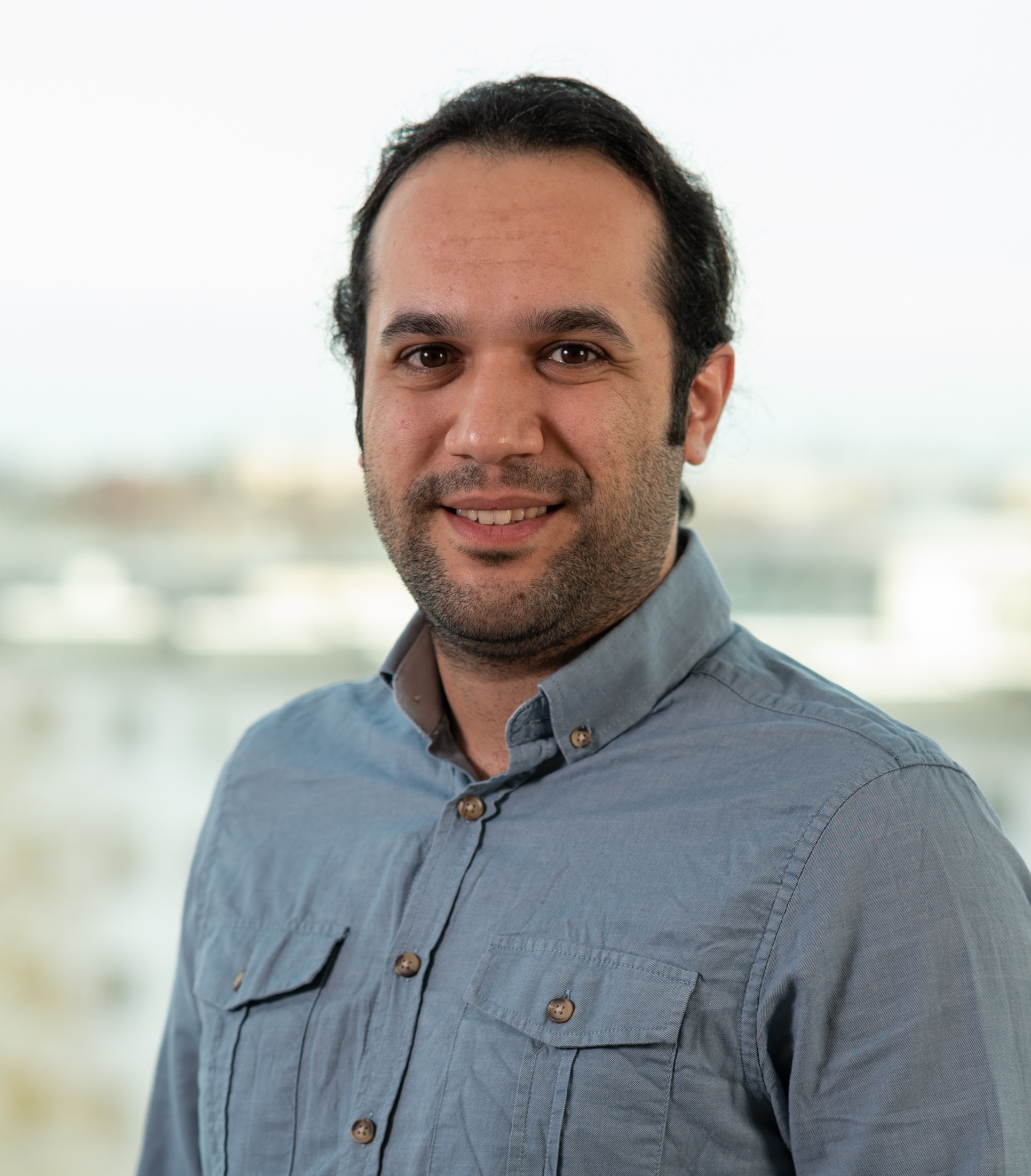}}]%
{
Mohammad Hasan Ahmadilivani} is a researcher at the Centre for Dependable Computing Systems, Computer Systems Department at Tallinn University of Technology (Taltech), Estonia. He received his Ph.D. at Taltech in 2025 and his M.Sc. in Computer Architecture Systems from the University of Tehran, Iran, in 2020. His research focuses on developing methods to measure and enhance the hardware reliability of Deep Neural Network (DNN) accelerators. His research interests include the safety of DL systems, robust ML and trustworthy AI, and efficient cross-layer fault tolerance techniques for edge AI applications. He has co-authored more than 20 scientific publications and served as a Local Organization Chair at the 30th IEEE European Test Symposium 2025.
\end{IEEEbiography}

\vspace{-8mm}

\begin{IEEEbiography}[{\includegraphics[width=1in,height=1.25in,clip,keepaspectratio]{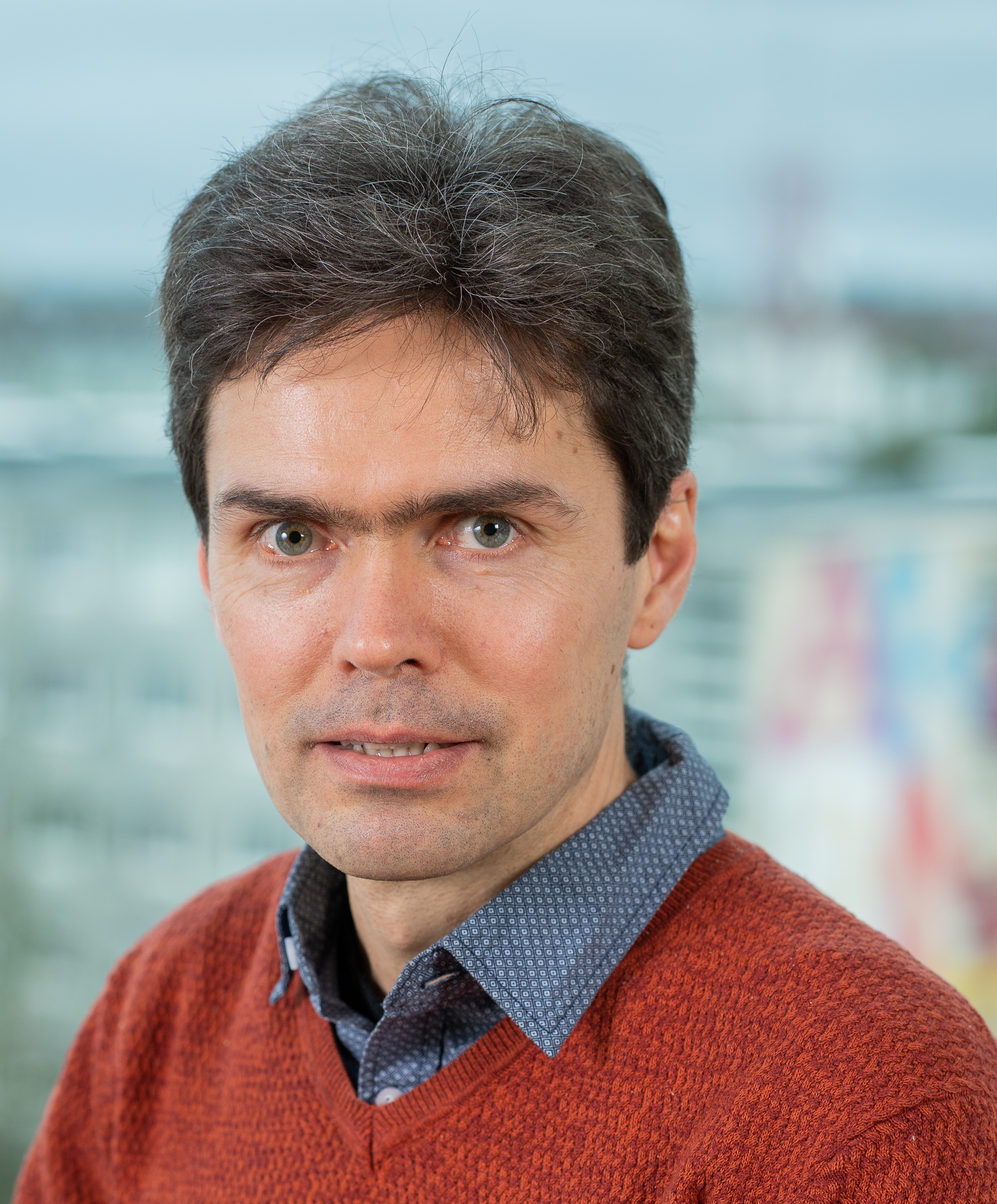}}]%
{
Jaan Raik} is a Full Professor at the Department of Computer Systems and Head of the Center for Dependable Computing Systems at the Tallinn University of Technology (Taltech), Estonia. He received his M.Sc. and Ph.D. degrees from Taltech in 1997 and in 2001, respectively. His research interests cover a wide area in electrical engineering and computer science domains including reliability of deep learning, hardware test, functional verification, fault-tolerance and security as well as emerging computer architectures. He has co-authored more than 400 scientific publications. He is a member of IEEE Computer Society, HiPEAC and of Steering/Program Committees of numerous conferences within his field. He served as the General Chair to IEEE European Test Symposium ’25, ’20, IFIP/IEEE VLSI-SoC ’16, DDECS ’12), Vice General Chair IEEE European Test Symposium ’24, DDECS ’13 and Program Co-Chair DDECS ’23, ’15, CDN-Live ’16 conferences. He was awarded the Global Digital Governance Fellowship at Stanford (2022) and HiPEAC Paper Award (2020).
\end{IEEEbiography}

\vspace{-8mm}

\begin{IEEEbiography}[{\includegraphics[width=1in,height=1.25in,clip,keepaspectratio]{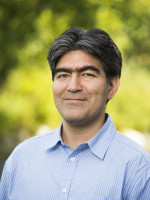}}]%
{
Masoud Daneshtalab} is a full professor at Mälardalen University in Sweden and adjunct professor at TalTech in Estonia. He is the director of the deep learning and heterogenous system (DeepHERO) lab. He is on the Euromicro board of directors and is an associate editor of the MICPRO journal. His research interests encompass algorithm-hardware co-design, embedded-friendly and reliable AI, and interconnection networks. He has authored over 220 journal and conference papers and has developed open-source tools that enhance AI reliability and performance, especially in safety-critical systems.
\end{IEEEbiography}

\vspace{-8mm}

\begin{IEEEbiography}[{\includegraphics[width=1in,height=1.25in,clip,keepaspectratio]{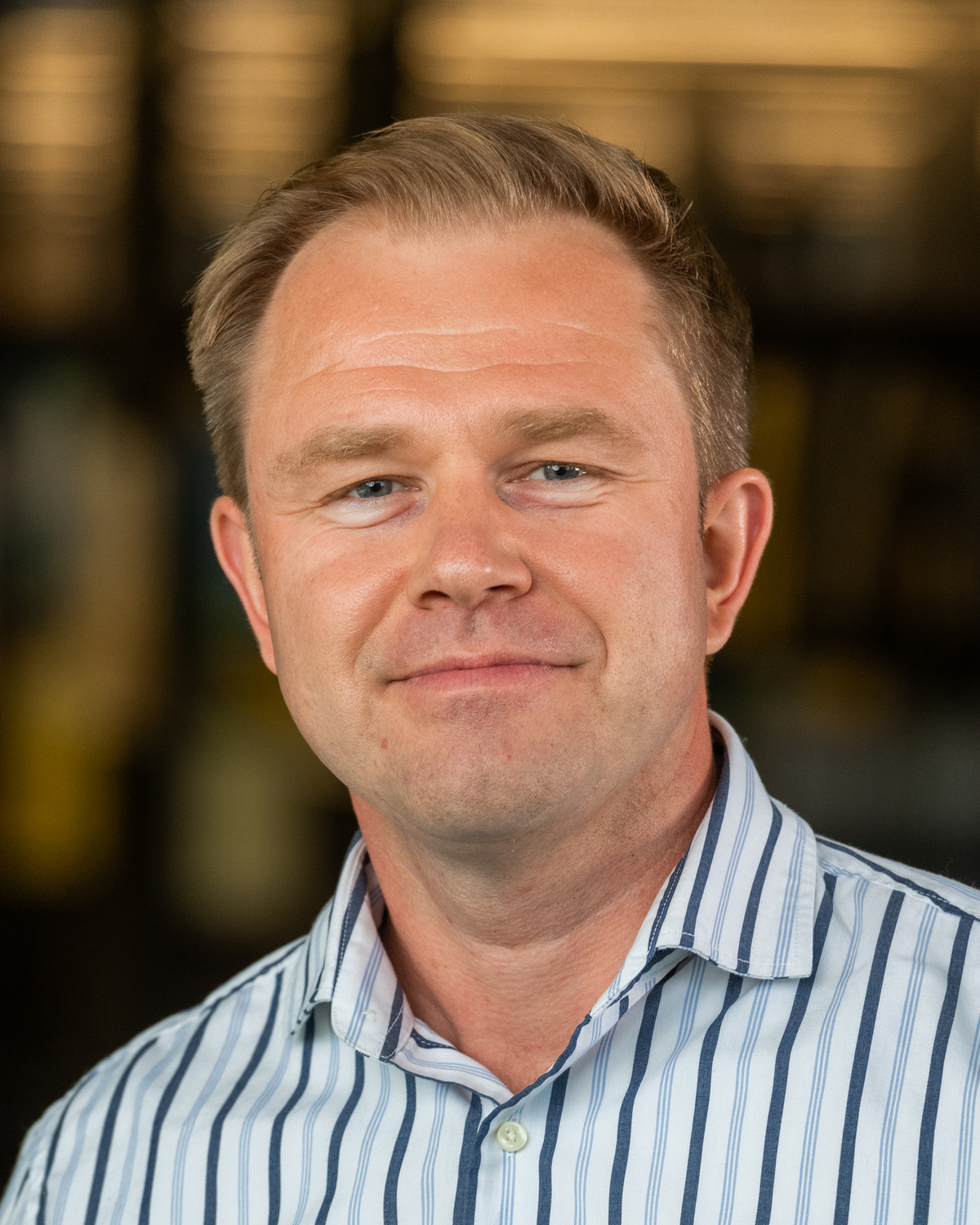}}]%
{
Maksim Jenihhin} is a tenured associate professor of computing systems reliability and Head of the research group Trustworthy and Efficient Computing Hardware (TECH) at the Tallinn University of Technology, Estonia. He received his PhD degree in Computer Engineering from the same university in 2008. His research interests include reliable and efficient hardware for AI acceleration, methodologies and EDA tools for hardware design, verification and security, as well as nanoelectronics reliability and manufacturing test topics. He has published more than 170 research papers, supervised several PhD students and postdocs and served on executive and program committees for numerous IEEE conferences (DATE, ETS, DDECS, LATS, NORCAS, etc.). Prof. Jenihhin coordinates European collaborative research projects HORIZON MSCA DN “TIRAMISU” (2024), HORIZON TWINN “TAICHIP” (2024) and national ones about energy efficiency and reliability of edge-AI chips and cross-layer self-health awareness of autonomous systems.
\end{IEEEbiography}
\vfill

\end{document}

%% file: sections/1-introduction.tex
\section{Introduction}  \label{sec:intro}

In recent years, the rapid advancement of Artificial Intelligence (AI), in particular of Machine Learning (ML), has provided a unique opportunity for automation, leading to AI exploitation in safety-critical applications \cite{wang2022artificial,rech2024artificial}. Deep Neural Networks (DNNs) are dominantly employed in ML for complex tasks such as image classification and object detection that have applications in the safety-critical domain \cite{athavale2020ai,moghaddasi2023dependable}. According to recent regulations such as the European Union's AI ACT \cite{AIACT} and the US National AI Initiative Act \cite{USACT}, AI systems that may cause adverse impacts on people’s safety are considered high-risk and must be rigorously assessed before deployment. This reflects the importance of early-stage safety risk assessment for AI deployment in safety-critical applications.

\begin{figure}[t!]
     \includegraphics[width=0.35\textwidth]{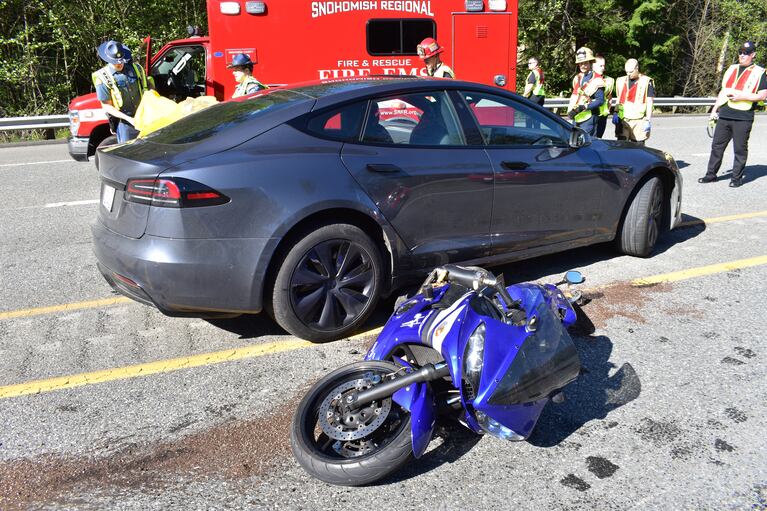}
 \centering
 \caption {A real-world example of an autonomous vehicle failure, leading to a fatal accident, source: \url{https://shorturl.at/QnOQq}. } \vspace{-6mm}
 \label{fig:example}
 \end{figure}

One of the primary concerns in safety-critical applications is ensuring the \emph{hardware reliability} of the underlying computational platforms concerning random hardware faults {\cite{rausand2014reliability}}.
In the context of DNN accelerators, hardware reliability refers to the probability that a DNN accelerator performs as intended during the deployment, despite potential faults in hardware {\cite{ahmadilivani2023deepvigor}}. Reliability assessment typically assumes the DL model and input data are error-free, focusing mainly on the impact of hardware faults on the application {\cite{bolchini2024resilience}}. Resilience analysis is the process of studying the impact of faults on the application's results {\cite{ahmadilivani2024systematic}}. Such faults appear due to soft errors, electromagnetic, temperature variation, aging, etc., that can perturb a DNN's parameters or activation values during runtime, potentially leading to catastrophic consequences. As an example, a hardware fault may result in failing the detection of objects on the road and result in a fatal accident, as shown in Fig. {\ref{fig:example}}.

With the technology miniaturization, the rate at which hardware faults occur is continuously rising in logic and memory circuits since the sensitivity of transistors increases against soft errors, temperature variations, etc. \cite{baumann2002impact,henkel2013reliable,safari2022survey}. 
This trend significantly impacts the reliability of DNN accelerators, especially as they are increasingly adopted in applications such as perception modules in autonomous vehicles \cite{yatbaz2023introspection}. Therefore, assessing the hardware reliability of DNNs during the design and development phases is necessary, not only to meet safety certification requirements but also to proactively address potential reliability issues before deployment.

%As DNNs are penetrating such applications, e.g., perception in automotive \cite{yatbaz2023introspection}, their hardware reliability assessment is necessary during the design and development phase, not only for risk certification but also for eliminating reliability issues as early as possible. 

%\vspace{-2mm}
\begin{figure}[b!]
    \includegraphics[width=0.45\textwidth]{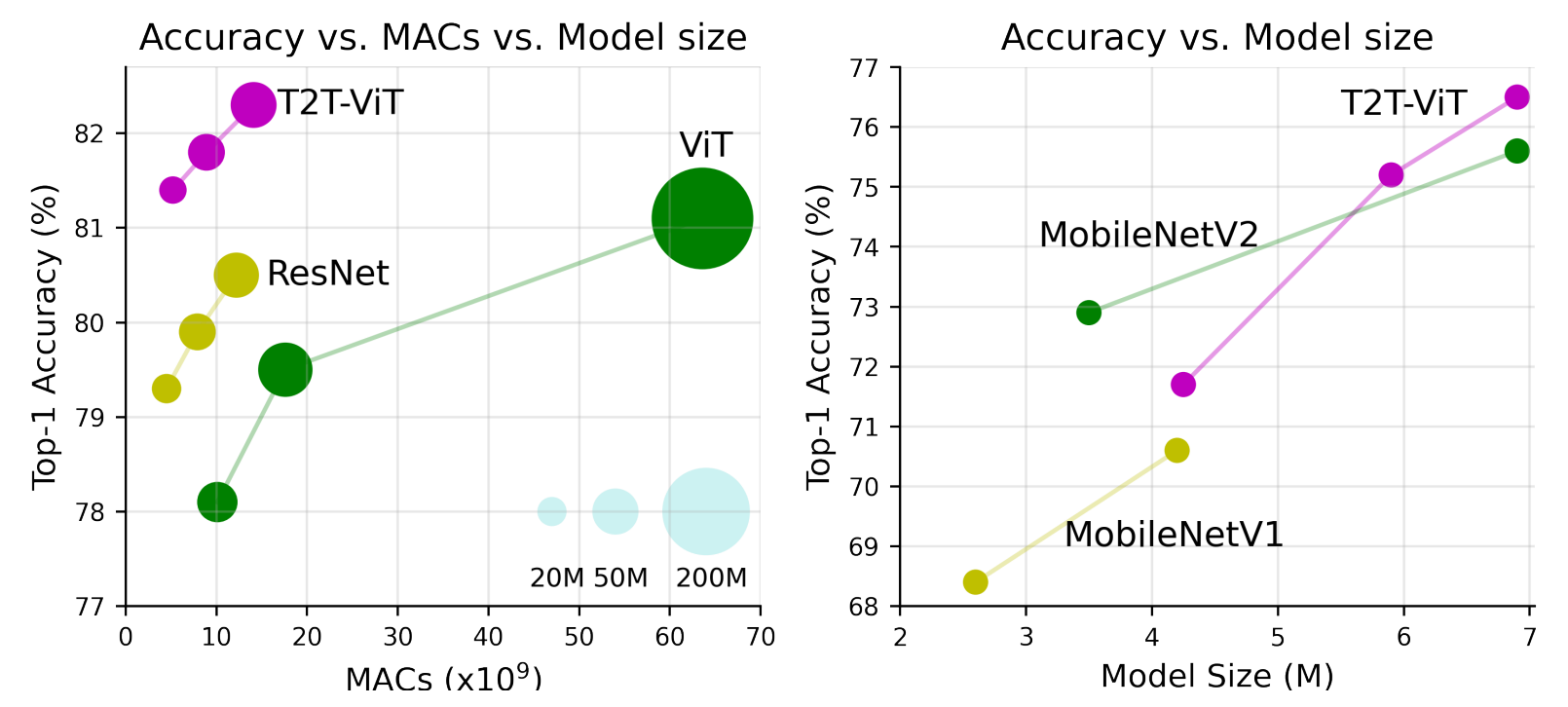}
\centering
\caption {Growing size of emerging DNN models regarding computation and memory requirement \cite{yuan2021tokens}. \vspace{-3mm}
}
\label{fig:cnn-comp}
\end{figure}

On the other hand, the size of DNNs in terms of memory footprint and number of operations is rapidly growing to achieve higher accuracy \cite{hussain2022design,desislavov2023trends}. Fig. \ref{fig:cnn-comp} presents the growing model size in terms of the number of MAC operations and parameters for emerging Convolutional Neural Networks (CNNs) and Vision Transformers (ViT) \cite{yuan2021tokens}. This trend necessitates leveraging more complex and performant computing systems such as GPUs, TPUs, etc. \cite{mohaidat2024survey}. The fact that these complex systems are prone to hardware faults \cite{ibrahim2020soft,dos2023understanding,garrett2024soft} leads to a necessary and challenging task of their hardware reliability assessment prior to deployment in safety-critical applications.

% \begin{figure}[h!]
%     \includegraphics[width=0.4\textwidth]{images/example-ai.jpeg}
% \centering
% \caption {An example of hardware faults' effect on an autonomous vehicle, generated by \url{https://getimg.ai}.}
% \label{fig:example}
% \end{figure}

To ensure a reliable DNN deployment, the first step is to extensively evaluate the performance and functionality of pre-trained DNNs against hardware faults.
In this regard, several research works have been carried out to assess and enhance DNNs' hardware reliability with respect to hardware faults listed in the related surveys \cite{ibrahim2020soft,mittal2020survey,shafique2020robust,su2023testability,ahmadilivani2024systematic}.
To achieve this objective, three main approaches are identified throughout the state-of-the-art, in a prior systematic literature review \cite{ahmadilivani2024systematic}: 1) Fault Injection (FI) methods in which faults are injected into the target system and simulated, 2) Analytical methods where faults effects are mathematically analyzed, and 3) Hybrid methods that combine FI and analytical methods. 

It was observed that the majority of existing research works adopt FI (based on simulation, emulation, or irradiation). In FI experiments, faults can be accurately modeled, and the behavior of a system is evaluated in the presence of faults. However, given the growing size of emerging DNNs and their accelerators {\cite{mohaidat2024survey,canziani2016analysis}}, obtaining a fully precise evaluation by FI requires billions of simulations, which takes weeks of execution, making it unfeasible and impractical.

Throughout the literature, multiple research works are presented to significantly reduce the simulation complexity of FI experiments for DNNs while maintaining statistical accuracy. It includes software simulation \cite{chen2019binfi,ruospo2023assessing, PytorchFI2020, weng2024fkeras}, or hardware emulation \cite{khoshavi2020fiji,taheri2023appraiser,taheri2024saffira}. Nonetheless, any FI experiment requires a certain level of statistical confidence, which is obtained by a considerable number of repetitions and a multitude of fault locations \cite{ruospo2023assessing}. Therefore, FI-based approaches for reliability assessment are inherently non-scalable and they take days and weeks using powerful computing resources such as GPUs.

On the other hand, analytical approaches are presented to tackle the scalability issue. Nonetheless, they cannot provide reliability-oriented metrics, including Vulnerability Factors (VFs), which are an essential metric for reliability evaluation. VF is the probability that a fault in a particular structure will result in an error. VF is in line with the probabilistic nature of reliability evaluation {\cite{mukherjee2003systematic}}, and it is widely used for reliability analysis of DNNs {\cite{ahmadilivani2024systematic}}. However, existing analytical methods attempt to achieve a relative contribution of neurons/weights to construct the classification outputs, and do not represent the impact of faults on the outputs of DNNs in a probabilistic way. Therefore, the accuracy of analytical methods is not comparable to that of FI-based methods.

To tackle the existing issues in the literature, a prior semi-analytical method named DeepVigor \cite{ahmadilivani2023deepvigor} is proposed as an alternative to simulation-based FI. DeepVigor employs analytical fault propagation combined with simulation-based computations, categorized as "semi-analytical", providing accurate VF for bits, neurons, and layers of CNNs. It is demonstrated that DeepVigor is faster than FI by utilizing a fault propagation analytical method. However, it analyzes all neurons in a CNN as well as exploits a huge search space of real numbers to obtain vulnerability values, leading to a scalability issue for large and deep CNNs. 

In this paper, we propose a scalable, fast, and accurate alternative to FI named DeepVigor+, which provides VF for layers and models of CNNs. DeepVigor+ takes advantage of a theoretical fault propagation analysis within neurons and entire DNNs to derive VFs in an optimized way. To the best of our knowledge, DeepVigor+ is the first scalable semi-analytical alternative to FI with a comparable accuracy for resilience analysis of DNNs in the literature. The contributions of this paper are as follows:

\begin{itemize}
    \item Introducing a scalable, fast, and accurate resilience analysis method for CNNs called DeepVigor+ for deriving VFs for convolutional layers by analyzing both parameters and activations. DeepVigor+ is built on a theoretical ground to enable an optimal error propagation analysis. Moreover, it conducts a novel statistical approach based on stratified sampling, unleashing a fast and accurate resilience analysis for large and deep CNNs,
    
    \item Extensively demonstrating the effectiveness of DeepVigor+ by examining its error compared to FI, leading to meeting a $1\%$ mean absolute error in obtaining VF. Furthermore, its scalability is evidenced against statistical FI, resulting in $14.9$ to $26.9$ times fewer simulations than the best state-of-the-art statistical FI approach (i.e., data-aware statistical FI {\cite{ruospo2023assessing}}). Derived VF by DeepVigor+ can provide the possibility of reliability comparison and visualization between and within various DNN models in a few minutes for deep and large DNNs.
\end{itemize}

DeepVigor+ is presented as an open-source tool (\url{https://github.com/mhahmadilivany/DeepVigor}), unlocking reliability analysis for emerging DNNs and enabling researchers to quickly assess their reliability and develop fault-tolerant DNNs. %To our knowledge, DeepVigor+ is the first analytical method with accuracy to FI, while significantly outperforming its simulation complexity.

Table \ref{tab:abbr} presents the abbreviations frequently used in the manuscript. In the rest of the paper, Section \ref{sec:related-works} highlights the gap by overviewing the related works on statistical FI as well as analytical methods. Section \ref{sec:method} presents the methodology of DeepVigor+. %leading to fast and accurate reliability analysis for CNNs. 
Section \ref{sec:setup} explains the implementation of DeepVigor+ and Section \ref{sec:experiments} presents the results. Section \ref{sec:discussion} presents the limitations and sheds light on the potential impacts of DeepVigor+ for reliability analysis of emerging DL systems, and finally, Section \ref{sec:conclusion} concludes the paper.

\begin{table}[t!]
\centering
\caption{Frequent abbreviations used in this manuscript.}
\label{tab:my-table}
%\resizebox{\textwidth}{!}{%
\begin{tabular}{ll}
\toprule\toprule
Abbreviation & Description                       \\ \toprule\toprule
DNN          & Deep Neural Network               \\ 
CNN          & Convolutional Neural Network      \\ 
FI           & Fault Injection                   \\ 
SFI          & Statistical Fault Injection       \\ 
VF           & Vulnerability Factor              \\ 
CVF          & Channel Vulnerability Factor      \\ 
LVF          & Layer Vulnerability Factor        \\ 
MVF          & Model Vulnerability Factor        \\ 
OFMap        & Output Feature Map                \\ 
IFMap        & Input Feature Map                 \\ 
VVSS         & Vulnerability Values Search Space \\ 
EDM          & Error Distribution Map            \\ 
CVV          & Candidate Vulnerability Value     \\ 
VVR          & Vulnerability Value Range         \\ 
MAE          & Mean Absolute Error               \\ 
CSR          & Channel Sampling Ratio            \\
\bottomrule
\end{tabular}%
%}
\vspace{-5mm}
\label{tab:abbr}
\end{table}

%% file: sections/2-related-works.tex
\section{Related Works: Resilience Analysis Methods for DNNs}  \label{sec:related-works}

With the fast growth of emerging DNNs' size regarding their memory and computation requirements, their reliability analysis becomes prohibitively complex, time-consuming, and resource-hungry. In a recent literature review \cite{ahmadilivani2024systematic}, it has been observed that the majority of papers studying the reliability of DNNs exploit a Fault Injection (FI) approach. Software-level simulation FI is widely adopted in reliability analysis due to its low design time and fast execution.

Multiple open-source tools for software-level simulation-based reliability analysis are proposed in the literature, and the most adopted ones include PytorchFI \cite{PytorchFI2020} and TensorFI \cite{chen2020tensorfi,laskar2022tensorfi+}, enabling FI simulations in PyTorch and TensorFlow, respectively. These software-level FI tools support various fault models and provide different metrics for the resilience analysis of DNNs regarding parameters and activations. 

Any FI experiment requires multiple simulations while a random set of parameters/activations are being faulty and propagated through the forward execution of a DNN. Then, the erroneous outputs are observed, and reliability metrics are acquired. The number of simulations affects the accuracy of the obtained metrics. In exhaustive FI, all the bits in the fault space are to be flipped and simulated individually, leading to a huge simulation space, which is impractical since DNNs possess millions of parameters and activations. Statistical Fault Injection (SFI) attempts to reduce the number of simulations based on statistical analysis, while guaranteeing a certain maximum error margin. 

For conventional computing hardware, the number of simulations in an SFI experiment is specified by Eq. \eqref{eq:SFI}, where \textit{n} is the number of samples, \textit{N} is the total fault space, \textit{e} is the error, and \textit{t} is determined based on the expected confidence-level \cite{leveugle2009statistical}.

\begin{equation}
    n = \frac{N}{1 + e^2 \times \frac{N-1}{t^2 \times p(p-1)}}
    \label{eq:SFI}
\end{equation}

However, it is shown that applying Eq. \eqref{eq:SFI} to the entire DNN does not result in statistically accurate results \cite{ruospo2023assessing}. Authors in \cite{ruospo2023assessing} introduce multiple methods to improve the accuracy of SFI as well as to reduce the number of simulations. Accordingly, layer-wise SFI, data-unaware SFI, and data-aware SFI are presented, which generally apply Eq. \eqref{eq:SFI} to each layer separately. Data-unaware SFI and data-aware SFI methods consider FI at the bit level to improve the statistical experiments. In data-unaware \textit{p = 0.5} for all bits in Eq. \eqref{eq:SFI}, whereas in data-aware SFI requires a pre-analysis to specify the \textit{p} for each bit position in the data representation. 

Yet, SFI requires hundreds of thousands of simulations, leading to a significantly long simulation time, in the order of weeks, even with leveraging GPUs. Furthermore, the growth of emerging DNNs necessitates an increasing number of simulations. Therefore, reliability analysis requires a paradigm shift in resilience analysis. Analytical methods for reliability analysis of DNNs are proposed to address the scalability issue.

Throughout the literature, multiple techniques are proposed to obtain the resilience of CNNs based on non-FI approaches, called analytical methods \cite{ahmadilivani2024systematic}. 
Layer-wise Relevance Propagation (LRP) is proposed to derive the contribution of neurons to the classification output of CNNs, expressing their relative criticality in the presence of faults and errors \cite{schorn2018accurate}. However, this notion of criticality primarily captures the potential impact of a neuron's malfunction on the output, rather than providing a probabilistic measure of classification errors due to faults. As such, it does not directly reflect the neuron's reliability.

The gradient-based analysis is exploited in \cite{wang2024enhancing,amarnath2022soft,amarnath2023error,sabih2021fault,mahmoud2020hardnn} to obtain the resilience of CNNs against faults. Nonetheless, gradients do not represent the impact of faults on classification and do not result in a probabilistic outcome of the faults' effect on the outputs. Also, faults may change the values significantly, while gradients represent small variations in erroneous values.

Thus, the existing analytical approaches do not result in accurate metrics for the reliability analysis of DNNs. DeepVigor was proposed to address this gap in the literature to provide fast and accurate metrics for the resilience of CNNs against faults \cite{ahmadilivani2023deepvigor}. It provides a Vulnerability Factor (VF) for bits, neurons, and layers in CNNs by acquiring vulnerability values for each neuron (i.e., output feature maps). Vulnerability values represent how much a fault should change the golden value of a neuron's output to change the CNN's golden classification. 

Although DeepVigor obtains VFs with a high accuracy (i.e., $<0.1\%$) faster than FI, there are a few drawbacks in its method, which make it non-scalable for deep and large CNNs. The drawbacks of DeepVigor and the ways DeepVigor+ addresses them are as follows:

\begin{itemize}
    \item To obtain VFs, DeepVigor first derives vulnerability value ranges with a huge search space based on the real numbers. However, 32-bit floating-point values do not represent all possible real numbers. To address this, DeepVigor+ builds a theoretical ground for optimal error propagation and then constructs a vulnerability value search space that includes all possible errors produced due to bitflips in 32-bit floating-point data representation. 
    
    \item To derive VFs, DeepVigor performs a bitflip mapping step in its operation, which conducts an exhaustive bitflip for each neuron to map the produced errors to vulnerability value ranges. This step is another reason that hinders DeepVigor from scaling. To tackle it, in this paper, we propose a mathematical approach to compute VFs from vulnerability value ranges with minimal computations based on the distribution of errors. 
    
    \item DeepVigor analyzes all neurons in a CNN, i.e., Output Feature Maps (OFMaps), in layers. This exhaustive analysis hinders a fast resilience analysis for large and deep CNNs and makes it impractical. To address this, we propose a stratified sampling method in DeepVigor+ to address the scalability of resilience analysis, while meeting the statistical requirements (i.e., $<1\%$). 
    
    \item DeepVigor only analyzes the activations to achieve VF and excludes analyzing the fault resilience of weights. To address this gap, we analyze both activations and weights in DeepVigor+ and obtain the Model Vulnerability Factor (MVF) for an entire CNN. 
\end{itemize}

%Although DeepVigor is demonstrated to be faster than FI due to its fault propagation analysis method, it requires a high execution time to obtain VFs. The reason is that DeepVigor attempts to analyze all neurons in a CNN as well as to find the vulnerability values in a huge space of numbers. 

%This paper proposes DeepVigor+, a novel method to quickly obtain VF metrics for layers and DNN models, addressing both scalability and accuracy of fault resilience analysis in the literature. DeepVigor+ employs a mathematical error propagation analysis in neurons and filters, assuming that a single fault might happen either at its inputs or weights, thus leading to effectively shrinking the search space for vulnerability values. Moreover, DeepVigor+ proposes a stratified sampling to further accelerate the process of resilience analysis without a tangible analysis accuracy mitigation, leading to obtaining VF in a few minutes, even for deep and large CNNs.

%% file: sections/3-method.tex
\section{Methodology: Scalable and Accurate Resilience Analysis for DNNs by DeepVigor+}   \label{sec:method}

In this section, the proposed resilience analysis method for DNNs, DeepVigor+, is presented. First, the fault model and its effect on 32-bit floating point values are described. Then, the mathematical model for fault propagation in DNNs is explained, which forms the basis for the DeepVigor+ analysis.

\subsection{Fault Model}

Transistor miniaturization leads to more efficient computing systems, whereas its reliability side effects are becoming increasingly profound. With transistor scaling, soft error rates increase significantly, posing reliability concerns, in particular in the deployment of safety-critical applications \cite{chatterjee2014impact,sullivan2021characterizing,yan2020single}. Single Event Upset (SEU) due to soft errors may affect memory cells and flip a bit in a stored value. SEU in DNN accelerators affects either the parameters of a DNN (i.e., weights) or its activations (i.e., layers' inputs), which are stored in memory elements such as registers or on-chip memories \cite{ibrahim2020soft,ahmadilivani2024systematic}. 

In this paper, we build the mathematical analysis based on the single-bit error model in weights and activations separately. We assume that a single parameter or a single input to a layer is faulty, which is propagated to the output of the corresponding layer. By mathematical analysis, we model the fault propagation to the output of neurons and CNNs, resulting in calculating VF for them. In reliability analysis, it is not feasible to exploit the multiple-bit fault model since it requires a huge fault space to consider all the $2^n$ combinations, where $n$ is the number of bits and each bit can be in one of the two states: fault-free, or faulty. On the other hand, it is demonstrated that tests achieving full fault coverage on the single-bit fault model cover a vast majority of multiple-bit faults as well {\cite{bushnell2004essentials}}. 
%Therefore, analyzing the reliability of DNNs based on a single-bit fault model is valid for obtaining the VF of models. This fault model is in line with the other works in the literature \cite{ahmadilivani2023deepvigor,chen2019binfi}.

\subsection{Theoretical Single Fault Analysis in 32-bit Floating-Point} \label{subsec:single-fault}

\begin{figure}[h!]
    \centering
    \includegraphics[width=.45\textwidth]{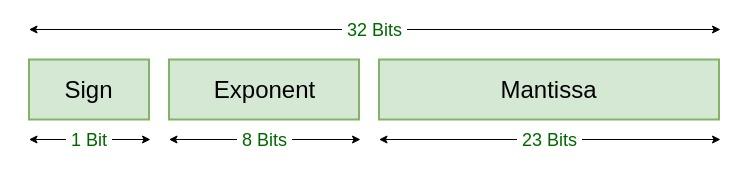}
    \vspace{-3mm}
    \caption{32-bit floating point IEEE-754 data representation.}
    \label{fig:32fp}
\end{figure}

%To model the behavior of faults for the analysis, we assume that at an inference, a single fault may influence the value of a single input activation to a neuron or a single parameter in a CNN. 
This subsection analyzes the effect of a single bitflip on the 32-bit floating-point data representation to comprehend how a value might change due to a single fault. The IEEE-754 32-bit floating-point data type is shown in Fig. \ref{fig:32fp}. It contains 1 sign bit, 8 exponent bits, and 23 mantissa bits, and represents a number based on Eq. \eqref{eq:fp32}.

\begin{equation}
\vspace{-2mm}
    number = (-1)^{sign} \times 2^{E-127} \times (1 + \sum_{i=0}^{i=23}{b_{23-i} \times 2^i})
    \label{eq:fp32}
\vspace{-3mm}
\end{equation}

To model single bitflips, we consider the following lemmas:

\begin{itemize}
    \item \textit{Lemma 1:} Any bitflip from 1 to 0 in a value decreases the value by $\epsilon$ while $\epsilon < 0$, and any bitflip from 0 to 1 increases it by $\epsilon$ while $\epsilon > 0$. 
    
    \item \textit{Lemma 2:} In the 32-bit floating-point data representation, an error ($\epsilon$) induced to a value ($x$) by a bitflip in bit $i$ can be represented as in Eq.~\eqref{eq:bitflip-value}, whether the bitflip is in sign bit ($2\times x$), exponent bits ($2^i \times x$), or mantissa bits ($2^i$), as stated in \cite{elliott2013quantifying,zhan2021improving}. 
    
    \begin{equation}
        \begin{cases}
            x_{faulty} = x + \epsilon \\
            \epsilon \in \{2 \times x, 2^{i-23} \times x, 2^i\}
        \end{cases}  
        \label{eq:bitflip-value}
    \end{equation}

    \item \textit{Lemma 3:} To unify the analysis of the error induced to a value by a single bitflip, we can approximate $\epsilon$ by representing it as a power of two, for each section of the data representation, as follows:
    \begin{enumerate}
        \item Sign: approximate $\epsilon$ as $\pm 2^{log_2(2\times x)}$.
        \item Exponent: approximate $\epsilon$ as $\pm 2^{i-23}$, assuming that $x$ is a small value.
        \item Mantissa: approximate $\epsilon$ as $\pm 2^{i}$.
    \end{enumerate}

    In all the cases, $\epsilon$ can be approximated as the nearest value of the power of two to the actual $\epsilon$. This approximation can lead to a unified representation for $\epsilon$. To that end, $\epsilon$ might be negative or positive (based on Lemma 1) and might be small or big (based on Lemma 2). Eventually, we can express it as a unified representation as shown in Eq. \eqref{eq:approx-epsilon}.
    \begin{equation}
        \epsilon \approx \pm 2^{\rho}; \rho \in \{\pm 1, \pm 2, \pm 3, ...\}
        \label{eq:approx-epsilon}
    \end{equation}

    \item \textit{Lemma 4:} When a faulty value is used in a multiplication operation in CNNs, the erroneous output can also be approximated. It has been observed that the parameters in CNNs are mainly distributed around 0 and in the range of $[-1, 1]$ \cite{huang2021rethinking}. In order to approximate the error of multiplying two values when one of them is faulty, we can analyze it based on Eq. \eqref{eq:faulty-mult}, where $x$ and $y$ are fault-free values and $x'$ is the faulty value after a bitflip in $x$.
    
    \begin{gather}
        let: \;\; x' = x + \epsilon \nonumber \\
        then: \;\; x' \times y = x \times y + \epsilon \times y \nonumber \\
        \Rightarrow x' \times y = x \times y + \delta
        \label{eq:faulty-mult}
    \end{gather}
    
    In Eq. \eqref{eq:faulty-mult}, when $x$ is a small value, $\delta \approx 0$. Since most values in CNNs' operations are close to $0$, the erroneous values in multiplications in a CNN can be approximated based on the unified error representation, as shown in Lemma 3 and Eq. \eqref{eq:approx-epsilon}.

\end{itemize}

\subsection{Single Bitflip Error Propagation in CNNs} \label{subsec:error-prop-cnn}

We analyze the single bitflip error propagation in CNNs, considering its effect on the values of numbers. Each neuron in a convolutional (CONV) layer operates as shown in Eq.~\eqref{eq:neuron}, where the Output Feature Map (OFMap) in $l$th layer and $k$th channel is obtained by the summation of multiplications between weights ($\hat{w}$) and Input Feature Map (IFMap, $\hat{x}$) plus bias ($b$). In CONV layers, $\hat{w}$ and $\hat{x}$ are 3-Dimensional (3D) $c_{in} \times n \times n$ arrays, where $c_{in}$ is the number of inputs channels to the layer, and $n$ is the kernel size. 

\begin{equation}
    OFMap^{l}_{k}(\hat{x}, \hat{w}, b) = (\sum_{i=0}^{c_{in}}\sum_{j=0}^n\sum_{k=0}^n{x_{ijk} \times w_{ijk}}) + b
    \label{eq:neuron}
\end{equation}

Here, we assume that a fault affects a single IFMap in a single neuron, thus, it produces a single erroneous OFMap. Supposing that a fault occurs in an IFMap $x_{ijk}$, represented as $x'_{ijk}$. The fault introduces an error $\epsilon$ to the fault-free value of $x_{ijk}$. Therefore, it can be expressed in Eq. \eqref{eq:faulty-ifmap}.

\begin{equation}
    x'_{ijk} = x_{ijk} + \epsilon
    \label{eq:faulty-ifmap}
\end{equation}

Hence, once a bitflip occurs in $\hat{x}$ in Eq.~\eqref{eq:neuron}, the partial multiplication is computed in Eq. \eqref{eq:faulty-mult-neuron}. In this equation, the term $x_{ijk} \times w_{ijk}$ represents the fault-free multiplication, and $\epsilon \times w_{ijk}$ is an added error to the output by a fault which can be represented as $\delta$.

\begin{gather}
    x'_{ijk} \times w_{ijk} = (x_{ijk} + \epsilon) \times w_{ijk} \nonumber\\
    = x_{ijk} \times w_{ijk} + \epsilon \times w_{ijk} \nonumber\\ 
    = x_{ijk} \times w_{ijk} + \delta
    \label{eq:faulty-mult-neuron}
\end{gather}

Consequently, the erroneous OFMap can be expressed in a way that it is the summation of the fault-free OFMap and $\delta$, while the single faulty IFMap in $\hat{x}$ can be in any index of the corresponding 3D array. Considering Lemma 4, the induced error at the neuron's OFMap can be expressed in Eq. \eqref{eq:faulty-ofm}. 

\begin{gather}
    OFMap^{l}_{k}(\hat{x}', \hat{w}, b) = OFMap^{l}_{k}(\hat{x}, \hat{w}, b) + \delta \nonumber \\
    \delta \approx \pm 2^{\rho}; \rho \in \{\pm 1, \pm 2, \pm 3, ...\}
    \label{eq:faulty-ofm}
\end{gather}

\begin{figure}[b!]
    \centering
    \includegraphics[width=.45\textwidth]{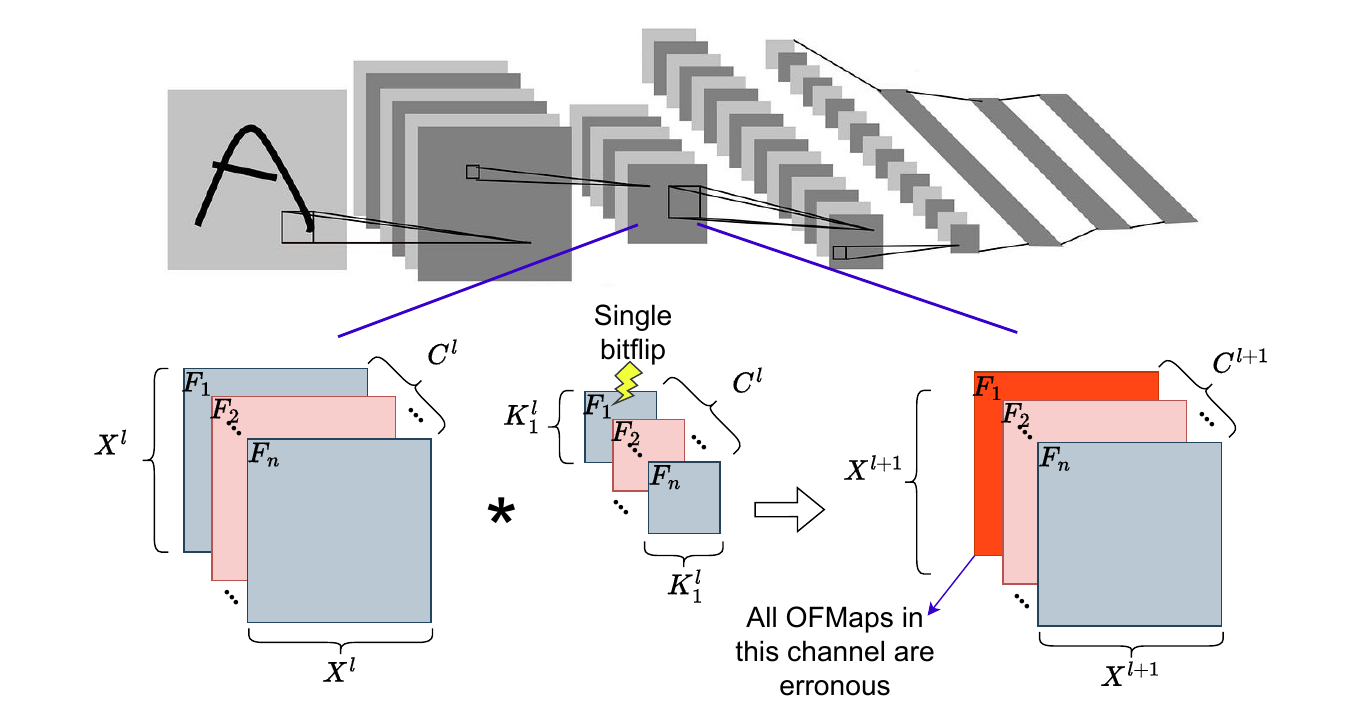}
    \caption{Fault propagation in the case of single bitflip in a filter. \vspace{-5mm}}
    \label{fig:fault-propagation}
\end{figure}

On the other hand, when a fault occurs in a weight, it has the same effect on a single neuron's output. However, the same faulty weight in the corresponding CONV layer is used by all neurons in the layer, as filters slide over all IFMaps. Therefore, it results in an output channel in which all OFMaps are erroneous, as shown in Fig. \ref{fig:fault-propagation}. Considering Lemma 4, the fault propagation for an output channel can be expressed in Eq. \eqref{eq:faulty-weight}.

\begin{gather}
    Out\_Channel^{l}_{k}(\hat{x}, \hat{w}, b) = \hat{x} * \hat{w} + b \nonumber\\
    Out\_Channel^{l}_{k}(\hat{x}, \hat{w}', b) = \hat{x} * \hat{w} + b  + \epsilon * \hat{x} \nonumber\\
    Out\_Channel^{l}_{k}(\hat{x}, \hat{w}', b) = Out\_Channel^{l}_{k}(\hat{x}, \hat{w}, b) + \delta \nonumber \\
    \delta \approx \pm 2^{\rho}; \rho \in \{\pm 1, \pm 2, \pm 3, ...\}
    \label{eq:faulty-weight}
\end{gather}

Based on the aforementioned theoretical analysis, to analyze the effect of single faults on CNNs, regardless of the fault occurrence in activations or weights, we need to identify an added value $\delta$ (as a power of two) at the output of the target neuron/channel where it changes the golden class of the CNN for each input image.

\begin{figure*}[t!]
    \centering
    \includegraphics[width=.85\textwidth]{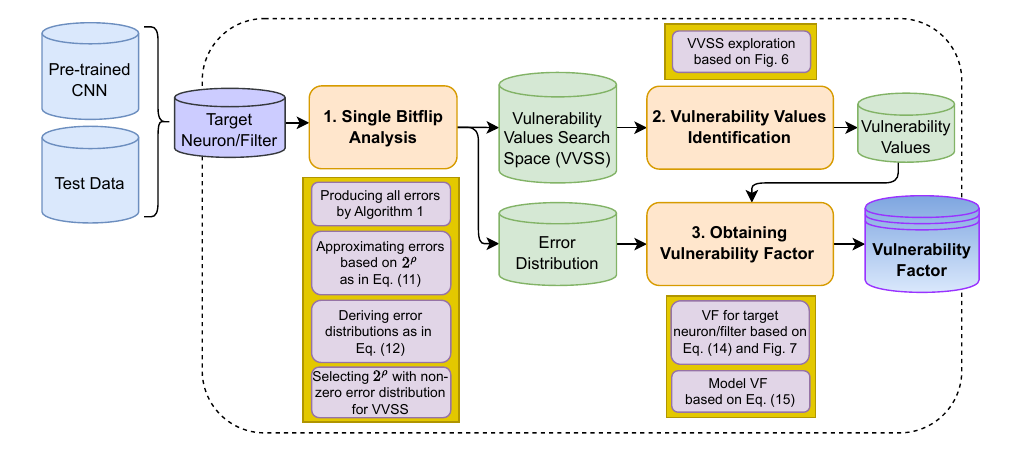}
    \caption{An overview of the DeepVigor+ approach for resilience analysis of CNNs.}
    \label{fig:deepvigor+method}
\end{figure*}

\subsection{DeepVigor+: Fault Resilience Analysis for DNNs} \label{subsec:deepvigor+}

In this subsection, the steps of DeepVigor+ are presented. Fig.~\ref{fig:deepvigor+method} illustrates the approach of the DeepVigor+ methodology for fault resilience analysis of CNNs against single faults. The method's objective is to provide the Vulnerability Factor (VF) for layers and the entire CNN model when a single fault happens in activations or weights.

The inputs of the method are a pre-trained CNN and test data in a dataset. During the analysis, an OFMap or output channel is targeted, and multiple steps are taken to provide the VF metrics: 

\begin{enumerate}
    \item Single Bitflip Analysis: constructs a search space for possible vulnerability values and produces a distribution of erroneous values based on the approximations in Eq. \eqref{eq:faulty-ofm} and \eqref{eq:faulty-weight},
    
    \item Vulnerability Values Identification: obtains vulnerability values for the target OFMap/channel,
    
    \item Obtaining Vulnerability Factor: exploits identified vulnerability values and error distribution to provide VF for the target neuron/channel.
\end{enumerate}

By obtaining the VF for the analyzed neurons and filters in a CNN, the VF for layers and the entire CNN can be derived. The details of each step are explained in the following. Note that each step presents a detailed description for analyzing a target neuron and then briefly links it to weights in filters.

%\vspace{3mm}
%\subsubsection{\textbf{Vulnerability Analysis for a Target Neuron:}}

\textbf{Step 1 - Single Bitflip Analysis:} %As mentioned above, single faults at the inputs of a neuron are considered. 
This step conducts bitflips in the input activations of the target neuron to construct Vulnerability Values Search Space (VVSS) and Error Distribution Map (EDM) for it. VVSS is a set of approximated produced output errors, as in Eq. {\eqref{eq:faulty-ofm}}, by single bitflips in any bit locations of the target neuron's inputs. EDM represents the distribution ratio of those approximated output errors. 

Obtaining VVSS and EDM requires performing a single bitflip for each input and multiplying it by its corresponding weight as shown in Eq. \eqref{eq:faulty-mult-neuron}. Nonetheless, it is an exhaustive operation with high time complexity. This complexity can be significantly reduced by using Algorithm \ref{alg:error-analysis-neuron}. In this algorithm, for each bitflip, we obtain all possible exact $\delta$ values produced at the output of the neuron for all inputs at once.

\begin{algorithm}[b!]
    \caption{One-Shot Bit-wise Error Analysis for a Neuron}
    \label{alg:error-analysis-neuron}
    \begin{algorithmic}[1]
\Require{Target neuron's inputs and weights as 3D matrices};
\Ensure{All possible errors at the output};
\Statex Assume: $\delta$ is the error added to each golden output; All values are in 32-bit floating-bit;
    \State binary\_representation = float32\_to\_binary(inputs);
    \For{$i \in [0, 31]$}:
        \State flip bit $i$ in binary\_representation;
        \State \brokenline{erroneous\_inputs = binary\_to\_float32(binary\_representation);}
        \State input\_errors = erroneous\_inputs - inputs;
        \State output\_errors\_list.append(input\_errors $\odot$ weights);
    \EndFor;
    \end{algorithmic}
\end{algorithm}

The faults in the input activations are independent for each input data. Hence, to produce all errors in a neural operation, we first flip the $i$th bit in all inputs (line 3) and then convert the binary representation to a value (line 4). Then, the difference of the \textit{erroneous\_inputs} with the fault-free input is computed (line 5), which represents all possible errors at inputs ($\epsilon$ in Eq. \eqref{eq:faulty-mult-neuron}) for all inputs added by a single bitflip in either of them. Then, a point-wise multiplication between the \textit{erroneous\_inputs} and \textit{weights} results in producing all possible errors ($\delta$ in Eq. \eqref{eq:faulty-ofm}) at the output (line 6).

Algorithm \ref{alg:error-analysis-neuron} produces all errors at the output of the target neuron at a single shot per bit. Thereafter, based on the presented theory in the subsections \ref{subsec:single-fault} and \ref{subsec:error-prop-cnn}, we generate the distribution of all produced errors as an Error Distribution Map (EDM) based on the Candidate Vulnerability Values (CVVs), which are a set of numbers as a power of two, shown in Eq. \eqref{eq:vulnerability-values}. Based on the experimental observations, we limit the CVV in the analysis between $-2^{10}$ and $2^{10}$ for large values. The error values between $-2^{-10}$ and $2^{-10}$ are discarded as their propagation effects on CNNs are negligible and masked.

\begin{equation}
    CVV = \{\pm 2^{\rho}\}; \rho \in \{0, \pm 1, \pm 2, \pm 3, \cdots, \pm 9, \pm 10 \}
    \label{eq:vulnerability-values}
\end{equation}

EDM contains the \textit{distribution ratio} of the existing values over the values in \textit{output\_errors\_list} in Algorithm \ref{alg:error-analysis-neuron}, between each consecutive value in CVV, as shown in Eq. \eqref{eq:dist-ratio}. EDM provides the ratio of error distribution for each candidate vulnerability value, demonstrating how much each CVV represents the produced errors at the output of the target neuron. Based on the EDM, each CVV whose \textit{distribution ratio} is not zero is added to the Vulnerability Value Search Space (VVSS). This means that VVSS contains a subset of CVV that are the approximated error values produced by single faults in the inputs. Therefore, VVSS only includes the CVVs with a non-zero error distribution ratio.

\begin{gather}
    \forall i \in CVV; distribution\_ratio_i = \nonumber \\
    \frac{count(CVV_{i-1} < output\_errors\_list < CVV_{i})}{count(output\_errors\_list)}
    \label{eq:dist-ratio}
\end{gather}

The process of single bitflip analysis for a target filter is similar to the one for a neuron. Nonetheless, in the case of a bitflip in a filter, all OFMaps in an output channel are affected (as described in subsection \ref{subsec:error-prop-cnn} and Eq. \eqref{eq:faulty-weight}). Therefore, this step is performed separately on every bit of all weights in the target filter, so that \textit{VVSS} and \textit{EDM} for the corresponding filter are obtained.

\textbf{Step 2: Vulnerability Values Identification:} 
This step exploits VVSS to identify the vulnerability values for the target neuron (i.e., OFMap) by exploring its constructed VVSS. %\hl{VVSS only contains the CVVs with a non-zero distribution ratio and represents output errors induced by a single fault occurring at the inputs of the target neuron.} 
The objective is to identify the \textit{maximum negative} and \textit{minimum positive} vulnerability values among the existing ones in VVSS for a neuron that changes the CNN from its golden classification for each input data. 

To explore the vulnerability values efficiently, we divide them into four different exploration spaces:
\begin{enumerate}
    \item $VVSS_{(-\infty, -1]}$: contains CVVs between $(-\infty, -1]$ that have a non-zero distribution ratio.
    \item $VVSS_{(-1, 0)}$: contains CVVs between $(-1, 0)$ that have a non-zero distribution ratio.
    \item $VVSS_{(0, 1)}$: contains CVVs between $(0, 1)$ that have a non-zero distribution ratio.
    \item $VVSS_{[1, +\infty)}$: contains CVVs between $[1, +\infty)$ that have a non-zero distribution ratio.
\end{enumerate}

\begin{figure*}[t!]
    \centering
    \includegraphics[width=.8\textwidth]{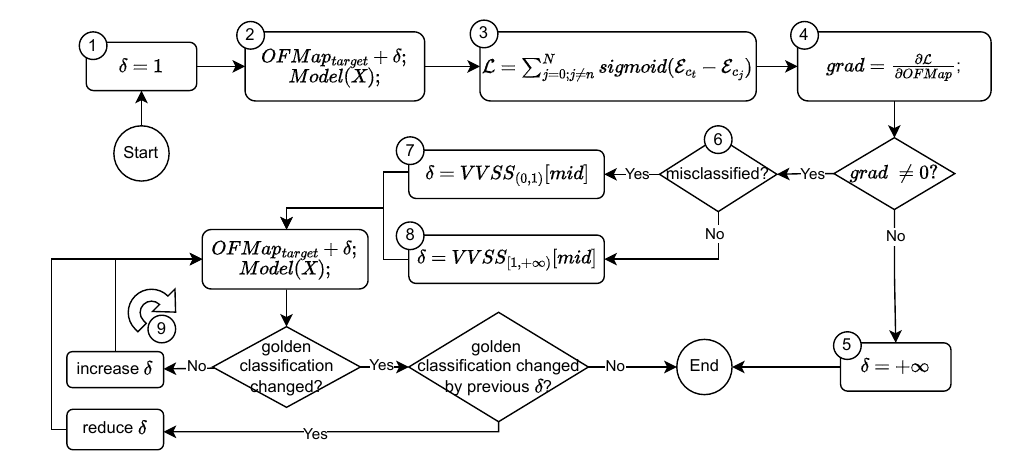}
    \caption{Vulnerability value identification for a target neuron with a single input data for positive errors.}
    \label{fig:DV+flowchart}
\end{figure*}

The flowchart in Fig~\ref{fig:DV+flowchart} illustrates the algorithm of vulnerability value identification for \textit{positive vulnerability values} for a single input data $\mathbf{X}$. In this flowchart, $\delta$ is the vulnerability value added to the target OFMap. First, $\delta$ equals 1 (box 1) is added to the target OFMap. Thereafter, the forward pass of the CNN is performed to obtain its output logits ($\mathcal{E}$) while a neuron is erroneous (box 2).

In box 3, a loss function $\mathcal{L}$ is calculated to obtain the effect of an added $\delta$ to the target OFMap on the output logits based on the summation of the differences between all output classes' logits from the golden class. In this loss function, $\mathcal{E}_{c_t}$ represents the erroneous output logit of the golden top class, and $\mathcal{E}_{c_j}$ is the erroneous output logit of any other class. 

In box 4, the gradient of loss function $\mathcal{L}$ w.r.t. the target OFMap is calculated to check if the CNN golden classification changes or not when the target OFMap is erroneous. If the gradient is $0$, faults producing positive deviation in the corresponding neuron do not lead to a change of golden classification. In this case (box 5), the neuron's vulnerability value is considered to be the biggest CVV assumed value for positive numbers (i.e., $2^{10}$) and the algorithm ends. If the gradient is not equal to $0$, a value can be found for $\delta$ in the target OFMap that changes the golden classification of the CNN. 

The rest of the algorithm attempts to find the minimum positive $\delta$ in VVSS. 
In box 6, the  CNN's classification results are examined when $\delta = 1$ for the target neuron. It initializes the VVSS subset to be explored in the next steps. If the golden classification changes due to $\delta = 1$, the vulnerability value is less than $1$, and we should explore $VVSS_{(0, 1)}$ (box 7). Otherwise, $VVSS_{[1, +\infty)}$ should be explored (box 8). Then, $\delta$ is set to the middle element of the corresponding VVSS subset, and if it does not change the golden classification, the next bigger CVV should be explored. This process continues until the first value that changes the golden classification is found (loop 9). The final positive vulnerability value ($\delta^{+}$) for the target neuron is the minimum value in positive VVSS that \textit{does not} change the golden class for each input. 

To identify the maximum negative vulnerability value ($\delta^{-}$) that does not change the golden classification, the same procedure is carried out by exploring negative values in VVSS. As a result, the Vulnerability Value Range (VVR) for the corresponding neuron is obtained and is expressed in Eq. \eqref{eq:vvr}. VVR represents all induced error values in the outputs of a neuron, leading to a change of classification. Note that since bitflips in 32-bit floating-point might lead to large values, we assumed that any value larger than $2^{10}$ and less than $2^{-10}$ is critical for CNNs; therefore, these values represent $+\infty$ and $-\infty$ respectively.

\begin{equation}
    VVR = (-\infty, \delta^{-}] \cup [\delta^{+}, +\infty)
    \label{eq:vvr}
\end{equation}

The vulnerability values' identification step for a target filter is similar to the one for a neuron. It is conducted similarly to the illustrated flowchart in Fig. \ref{fig:DV+flowchart} to obtain VVR for the corresponding output channel, resulting in VVR for each target channel in a CNN.

\textbf{Step 3: Obtaining Vulnerability Factors:} 
As mentioned, an Error Distribution Map (EDM) is obtained for the target neuron in step 1, representing the distribution ratio for each vulnerability value. On the other hand, the Vulnerability Value Range (VVR) is obtained in step 2. This step aims to map VVR to EDM to provide the VF for the target neuron, expressing the probability of changing the golden classification in the case of a single bitflip in its inputs. 

\begin{figure}[h!]
    \includegraphics[width=0.48\textwidth]{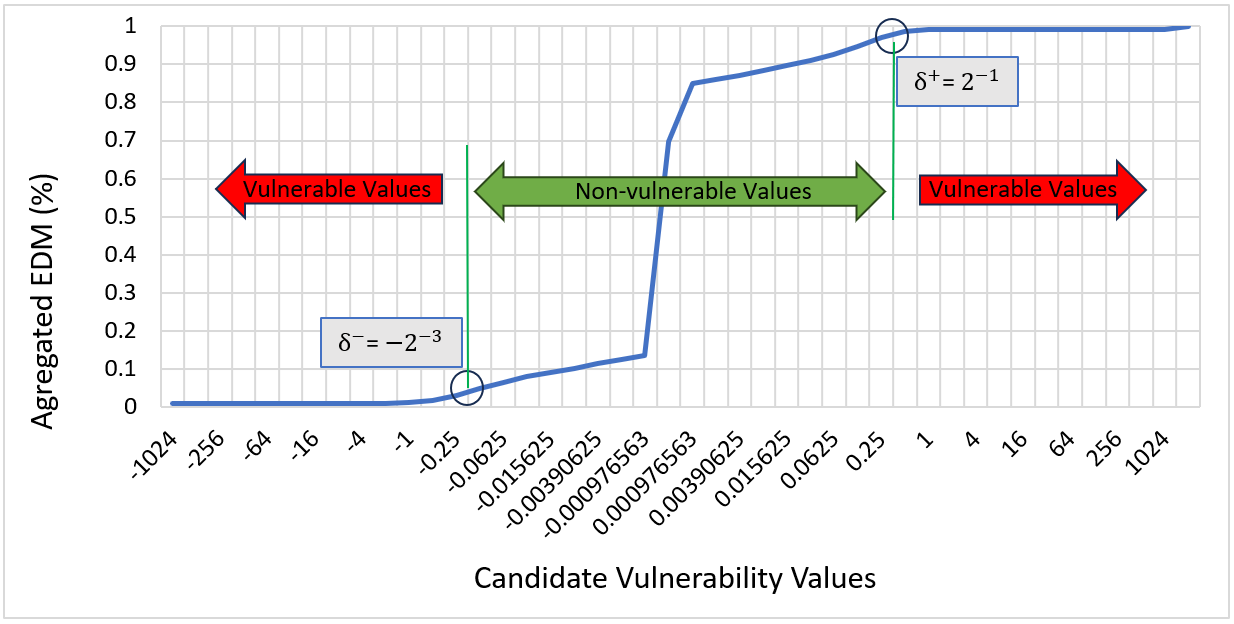}
\centering
\caption {Mapping obtained Vulnerability Value Range (VVR) to aggregated Error Distribution Map (EDM) for VF calculation. \vspace{-1mm}}
\label{fig:vf-calc}
\end{figure}

Fig. \ref{fig:vf-calc} depicts how VF can be obtained by mapping VVR to an aggregated EDM. The aggregated EDM represents the summation of the \textit{distribution ratio} for all values less than a CVV. As shown in Fig. \ref{fig:vf-calc}, all values between $(\delta^-, \delta^+)$ are non-vulnerable, meaning that if an error deviates the OFMap as much as any value in this range, it will not change the golden output. Otherwise, the error value is vulnerable, leading to a change in the golden classification. 

VF is obtained by subtracting the distribution ratio of the minimum positive vulnerability value ($\delta{+}$) and the maximum negative vulnerability value ($\delta^{-}$) as shown in Eq. \eqref{eq:vf}. This equation expresses the portion of all errors produced at the output that are critical for the CNN classification, which is equivalent to the probability that a fault changes the CNN's golden classification. 

\begin{gather}
    VF_{target\_neuron} = \nonumber \\ 
    1 - (distribution\_ratio_{\delta^{+}} - distribution\_ratio_{\delta^{-}})
    \label{eq:vf}
\end{gather}

The VF for a filter is obtained similarly. The EDM and VVR for the target filter are obtained in the previous steps. In this step, they are exploited to provide the VF for the target filter. Deriving the VF for individual neurons and filters leads to obtaining the VF for CONV layers and the entire CNN model. %Since the analysis is performed for neurons and filters separately, it can provide separate VF for layers and the entire model based on filters and neurons. 
%The Layer Vulnerability Factor (LVF) is the average of VF for all activations/filters, and the Model Vulnerability Factor (MVF) is the average of LVFs within it. 

To obtain the total MVF of the entire model %using the obtained detailed VFs based on the activations and filters analysis, 
Eq. \eqref{eq:mvf-total} is introduced, where $L$ is the total number of layers in a CNN, $N_l$ and $W_l$ are the total number of output activations and weights in layer $l$ respectively, $LVF_{neuron_l}$ represents the Layer Vulnerability Factor (LVF) based on the average VF of the neurons in layer $l$, and $LVF_{filter_l}$ is the average VF of the filters in layer $l$. This equation is a layer-wise weighted average of LVFs throughout the layers of a CNN that includes both activations and filters vulnerability analysis, leading to the MVF for the entire model. It represents the probability of changing the golden classification if a fault occurs either in activations or weights. 

\begin{gather}
    MVF_{total} = \nonumber \\
    \frac{\sum_{l=1}^L (\frac{N_l}{N_l + W_l} \times LVF_{neuron_l} + \frac{W_l}{N_l + W_l} \times LVF_{filter_l})}{L}
    \label{eq:mvf-total}
\end{gather}

%\vspace{3mm}
\subsection{\textbf{Stratified Random Sampling for Vulnerability Analysis:}}

The DeepVigor+ analysis is capable of analyzing all neurons and filters in a CNN. However, performing a complete analysis is computationally complex and prohibitively time-consuming, in particular for very deep CNNs. Inspired by statistical FI, random sampling is the main solution for reducing the number of simulations, while preserving acceptable accuracy. However, naive random sampling of neurons across the entire CNN model or within individual layers can result in poor representativeness and accuracy of the resilience analysis, since the total population is extremely large.

To address the scalability issue in resilience analysis for huge CNNs without sacrificing accuracy, we exploit the concept of stratified random sampling to reduce the population size for statistical DeepVigor+ analysis. Stratified random sampling is a method to enhance the statistical estimation accuracy by dividing the population into meaningful subgroups or \textit{strata}, and performing random sampling within them {\cite{singh1996stratified}}. This method first randomly selects a subset of strata and then samples from each selected stratum to ensure balanced and representative coverage.

In DeepVigor+, the analysis is performed in a layer-wise manner, while each layer consists of multiple channels. We consider each output channel in a layer as a \textit{stratum}, representing different features of the CNN. A portion of these output channels is randomly selected, specified by \textit{channel\_sampling\_ratio}. Within each stratum (i.e., output channel), a subset of neurons is then randomly selected for analysis, determined by the logarithm of the total neuron count in the channel.
This strategy ensures that the analysis remains computationally tractable while maintaining representative population coverage.
The total number of analyses conducted under this approach can be defined in Eq. \eqref{eq:analysis-count}.

\begin{gather}
    \#analyzed \; neurons = \sum_{l \in layers}\lceil channel\_sampling\_ratio \times \nonumber \\  
     out\_channel_{l} \rceil \times log_2(\#neurons^l_{channel})
     \label{eq:analysis-count}
\end{gather}

%% file: sections/4-setup.tex
\section{Experimental Setup} \label{sec:setup}

\subsection{DeepVigor+ Tool Implementation}

DeepVigor+ is fully implemented using Python and the PyTorch library. The source code of DeepVigor+ is fully open-source in \url{https://github.com/mhahmadilivany/DeepVigor} as a tool to enable researchers and engineers to adopt it for the resilience analysis of DNNs. The user can obtain the VF for the CONV layers of a target CNN based on analyzing neurons or weights by determining it through some inputs. The user can specify the following inputs for the analysis: 

\begin{itemize}
    \item model: The tool loads the pre-trained set of weights. By specifying the names of the model, it automatically loads the pretrained model. All the pretrained models from the open-source PyTorch CIFAR Models repository on GitHub\footnote{\url{https://github.com/chenyaofo/pytorch-cifar-models/tree/master}} and the PyTorch hub are integrated and supported within the tool. 
    \item dataset: The tool loads the test data from the dataset. It supports CIFAR-10, CIFAR-100, and ImageNet datasets.
    \item batch-size: Specifies the batch-size of data to be analyzed by the tool.
    \item run-method: The tool performs resilience analysis for neurons (OFMaps) or filters (weights). The user specifies the analysis method: 
    \begin{itemize}
        \item full-analysis-weight: Performs complete analysis on all weights of the CNN model,
        \item full-analysis-act: Performs complete analysis on all activations of the CNN model,
        \item sampling-analysis-weight: Performs sampling analysis on the logarithm number of weights per randomly selected channel based on \textit{channel sampling ratio},
        \item sampling-analysis-act: Performs sampling analysis on the logarithm number of activations per randomly selected channel based on \textit{channel sampling ratio},
    \end{itemize}
    \item Channel-sampling-ratio: In the case of \textit{sampling analysis}, this parameter specifies the percentage of the random channels in each CONV layer to be analyzed. 
\end{itemize}

With the determined inputs, the tool performs the analysis based on the presented method in Section {\ref{sec:method}} and outputs the Layer Vulnerability Factor (LVF) for each layer. The tool creates an individual directory for a pretrained model. In the directory, the tool creates a \textit{log} file and names it based on the specified \textit{model}, \textit{dataset}, \textit{run-mode}, and \textit{channel-sampling-ratio}, and saves the obtained VF results. Therefore, the Model Vulnerability Factor (MVF) for the entire CNN model can be calculated based on VF for activations and weights. In the paper, we perform the DeepVigor+ analysis on one batch of 100 images in the test set. Note that all experiments consider 32-bit floating-point data representation. 

\subsection{Validating DeepVigor+} \label{subsec:FI-setup}

We use Fault Injection (FI) to validate the results of DeepVigor+. To that end, we validate the VF for channels by analyzing filters of CNNs using complete DeepVigor+ and weights FI. In this regard, before an inference, a random 3D filter in a specified layer is selected. In the FI campaign, one bit of one weight in the target filter is flipped, considering 32-bit floating-point data representation, and the inference is performed on the same data as the DeepVigor+ analysis was conducted. The same FI process is carried out for all bits of weights in the target filter, resulting in the Channel Vulnerability Factor ($CVF_{FI}$) calculated by the ratio of changes in the golden classifications. $CVF_{FI}$ for each channel is compared with the $CVF_{weight}$ for the same channel in complete DeepVigor+, and their absolute errors are reported. In all DNNs under the experiment, $15\%$ of all channels throughout the CONV layers are passed through the FI campaign for validation. 

To show the accuracy of the results in stratified sampling in DeepVigor+, we compare the obtained VF for channels (CVF) and layers (LVF) in sampling DeepVigor+ vs. complete DeepVigor+ for activations and filters separately. The absolute difference between the obtained VFs is derived to show the VF estimation in sampling analysis. Different \textit{channel\_sampling\_ratio}s are explored including $5\%$, $10\%$, $15\%$, and $20\%$. Since neurons and channels are selected randomly in the stratified sampling, each sampling analysis is repeated $50$ times, and their maximum and Mean Absolute Error (MAE) are reported.

Furthermore, to show the efficiency of DeepVigor+ analysis, we present the number of simulations for DeepVigor+ and how it outperforms FI. First, we compare the number of simulations for DeepVigor+'s complete analysis against exhaustive FI. Also, we compare the number of simulations for sampling DeepVigor+ with state-of-the-art Statistical FI (SFI) methods for CNNs proposed in \cite{ruospo2023assessing}, considering layer-wise SFI, data-unaware SFI, and data-aware SFI. Finally, the paper demonstrates the execution time for the complete and sampling DeepVigor+ on an NVIDIA A100 GPU to showcase the efficiency of the proposed method. 

\subsection{CNNs Under Study}

In this paper, DeepVigor+ is executed and validated on six pre-trained DNNs using various datasets. CNNs under analysis include VGG-11 trained on CIFAR-10, VGG-16, ResNet-18-C, and MobileNetV2 trained on CIFAR-100, ResNet-18-I and ResNet-34 trained on ImageNet. The baseline accuracy, number of channels and neurons for each CNN are shown in Table \ref{tab:cnns}. All experiments in this paper are performed on an NVIDIA A100 GPU accompanied by an AMD EPYC 7742 64-core CPU.

\begin{table}[h!]
\small
\centering
\caption{The DNNs under study for DeepVigor+ validation.}
\resizebox{\columnwidth}{!}{%
\begin{tabular}{cccccc}\toprule\toprule
DNN     & Dataset   & \begin{tabular}[c]{@{}c@{}} Baseline \\ accuracy \end{tabular}  & \begin{tabular}[c]{@{}c@{}}\# of conv \\ layers \end{tabular} & \# of channels &  \# of neurons    \\ \toprule\toprule
VGG-11       &   CIFAR-10   &   92.52\%   &   8    &   2,752   &  232,448   \\ \toprule
VGG-16       &   CIFAR-100  &   66.97\%   &   13   &   4,224   &  185,344   \\ \toprule
ResNet-18-C  &   CIFAR-100  &   70.26\%   &   20   &   4,800   &  666,624   \\ \toprule
MobileNetV2  &   CIFAR-100  &   61.27\%   &   54   &   17,188  &  854,064   \\ \toprule
ResNet-18-I  &   ImageNet   &   69.19\%   &   20   &   4,800   &  2,182,656 \\ \toprule
ResNet-34    &   ImageNet   &   73.04\%   &   36   &   8,512   &  3,437,056  \\ \bottomrule
\end{tabular}
}
\label{tab:cnns}
\vspace{-5mm}
\end{table}

%% file: sections/5-results.tex
\section{Experimental Results} \label{sec:experiments}

\subsection{DeepVigor+ Accuracy with FI}

To analyze the accuracy of the obtained VFs by DeepVigor+, we perform exhaustive FI experiments on $15\%$ of randomly selected channels in all CNNs in Table \ref{tab:cnns} and compare the results with complete DeepVigor+, as described in subsection \ref{subsec:FI-setup}. Table \ref{tab:FI-validation} presents the MAE between $CVF_{FI}$ and $CVF_{weight}$ by the complete DeepVigor+'s filters analysis for each CNN. The acceptable mean error for a reliability analysis compared to exhaustive FI is $1\%$ \cite{ruospo2023assessing}.

The results in Table \ref{tab:FI-validation} indicate that the MAE by DeepVigor+ compared to exact FI throughout the CNNs is between $0.819\%$ to $0.978\%$, i.e., always less than $1\%$. These results demonstrate that DeepVigor+ is able to provide precise VFs and meets the expectation of an acceptable error with respect to exhaustive FI experiments.

\begin{table}[h!]
\small
\centering
\caption{MAE of VF for weight channels between complete DeepVigor+ and exhaustive FI for $15\%$ of the channels in CNNs.}
\resizebox{\columnwidth}{!}{%
\begin{tabular}{ccccccc} \toprule\toprule
CNN   
&  VGG-11  
&  VGG-16  
&  \begin{tabular}[c]{@{}c@{}} ResNet\\-18-C \end{tabular}   
&  \begin{tabular}[c]{@{}c@{}} Mobile-\\NetV2 \end{tabular}   
&  \begin{tabular}[c]{@{}c@{}} ResNet\\-18-I \end{tabular}  
&  \begin{tabular}[c]{@{}c@{}} ResNet\\-34 \end{tabular}  
\\ \toprule\toprule

\begin{tabular}[c]{@{}c@{}} Mean \\ Absolute \\ Error (\%) \end{tabular}       
&   \textbf{0.819}
&   \textbf{0.938}   
&   \textbf{0.933}  
&   \textbf{0.874}
&   \textbf{0.978}   
&   \textbf{0.878}   \\ \bottomrule
\end{tabular}
}
\label{tab:FI-validation}
\end{table}

The main sources of error in DeepVigor+ are as follows:
\begin{enumerate}
    \item As discussed in subsection \ref{subsec:error-prop-cnn}, DeepVigor+ approximates the error propagation in neurons based on the values of power of $2$. This error approximation introduces an error to the resilience analysis, which is also reflected in the VF calculation.
    
    \item DeepVigor+ assumes that bitflips resulting in big values are critical for CNNs, and the \textit{distribution ratio} for large values is always considered critical. However, in some cases, these values don't result in the change of the golden classification. This phenomenon is another reason for a minor difference between the VF by FI and DeepVigor+. 
\end{enumerate}

\subsection{Sampling Analysis vs. Complete Analysis}

\begin{table*}[h!]
\caption{Average absolute error analysis over 50 executions for sampling DeepVigor+ compared to complete analysis.}
\label{tab:sampling-error}
\resizebox{\textwidth}{!}{%
\begin{tabular}{|c|c|cccc||cccc|}
\hline
   &
   &
  \multicolumn{4}{c||}{Activations analysis} &
  \multicolumn{4}{c|}{Filters analysis} \\ \hline
CNN &
  \begin{tabular}[c]{@{}c@{}}channel \\ sampling \\ ratio\end{tabular} &
  \multicolumn{1}{c|}{\begin{tabular}[c]{@{}c@{}}MAE \\ $CVF_{neuron}$\end{tabular}} &
  \multicolumn{1}{c|}{\begin{tabular}[c]{@{}c@{}}Max error\\ $CVF_{neuron}$\end{tabular}} &
  \multicolumn{1}{c|}{\begin{tabular}[c]{@{}c@{}}MAE \\ $LVF_{neuron}$\end{tabular}} &
  \begin{tabular}[c]{@{}c@{}}Max error \\ $LVF_{neuron}$\end{tabular} &
  \multicolumn{1}{c|}{\begin{tabular}[c]{@{}c@{}}MAE \\ $CVF_{filter}$\end{tabular}} &
  \multicolumn{1}{c|}{\begin{tabular}[c]{@{}c@{}}Max error\\ $CVF_{filter}$\end{tabular}} &
  \multicolumn{1}{c|}{\begin{tabular}[c]{@{}c@{}}MAE \\ $LVF_{filter}$\end{tabular}} &
  \multicolumn{1}{c|}{\begin{tabular}[c]{@{}c@{}}Max error \\ $LVF_{filter}$ \end{tabular}} \\ \hline \hline

\multirow{4}{*}{VGG-11} &
  5\% &
  \multicolumn{1}{c|}{0.0004\%} &
  \multicolumn{1}{c|}{0.011\%} &
  \multicolumn{1}{c|}{0.0010\%} &
  0.012\% &
  \multicolumn{1}{c|}{0.064\%} &
  \multicolumn{1}{c|}{0.171\%} &
  \multicolumn{1}{c|}{0.019\%} &
  0.125\% \\ \cline{2-10} 
 &
  \cellcolor{yellow!30}\textbf{10\%} &
  \multicolumn{1}{c|}{\cellcolor{yellow!30}\textbf{0.0004\%}} &
  \multicolumn{1}{c|}{\cellcolor{yellow!30}\textbf{0.008\%}} &
  \multicolumn{1}{c|}{\cellcolor{yellow!30}\textbf{0.0008\%}} &
  \cellcolor{yellow!30}\textbf{0.008\%} &
  \multicolumn{1}{c|}{\cellcolor{yellow!30}\textbf{0.064\%}} &
  \multicolumn{1}{c|}{\cellcolor{yellow!30}\textbf{0.148\%}} &
  \multicolumn{1}{c|}{\cellcolor{yellow!30}\textbf{0.015\%}} &
  \cellcolor{yellow!30}\textbf{0.097\%} \\ \cline{2-10} 
 &
  15\% &
  \multicolumn{1}{c|}{0.0004\%} &
  \multicolumn{1}{c|}{0.004\%} &
  \multicolumn{1}{c|}{0.0007\%} &
  0.007\% &
  \multicolumn{1}{c|}{0.063\%} &
  \multicolumn{1}{c|}{0.130\%} &
  \multicolumn{1}{c|}{0.011\%} &
  0.083\% \\ \cline{2-10} 
 &
  20\% &
  \multicolumn{1}{c|}{0.0004\%} &
  \multicolumn{1}{c|}{0.006\%} &
  \multicolumn{1}{c|}{0.0007\%} &
  0.007\% &
  \multicolumn{1}{c|}{0.063\%} &
  \multicolumn{1}{c|}{0.128\%} &
  \multicolumn{1}{c|}{0.008\%} &
  0.076\% \\ \hlineB{2}

\multirow{4}{*}{VGG-16} &
  5\% &
  \multicolumn{1}{c|}{0.037\%} &
  \multicolumn{1}{c|}{0.141\%} &
  \multicolumn{1}{c|}{0.033\%} &
  0.250\% &
  \multicolumn{1}{c|}{0.045\%} &
  \multicolumn{1}{c|}{0.216\%} &
  \multicolumn{1}{c|}{0.017\%} &
  0.183\% \\ \cline{2-10} 
 &
  \cellcolor{yellow!30}\textbf{10\%} &
  \multicolumn{1}{c|}{\cellcolor{yellow!30}\textbf{0.038\%}} &
  \multicolumn{1}{c|}{\cellcolor{yellow!30}\textbf{0.144\%}} &
  \multicolumn{1}{c|}{\cellcolor{yellow!30}\textbf{0.022\%}} &
  \cellcolor{yellow!30}\textbf{0.188\%} &
  \multicolumn{1}{c|}{\cellcolor{yellow!30}\textbf{0.044\%}} &
  \multicolumn{1}{c|}{\cellcolor{yellow!30}\textbf{0.116\%}} &
  \multicolumn{1}{c|}{\cellcolor{yellow!30}\textbf{0.010\%}} &
  \cellcolor{yellow!30}\textbf{0.094\%} \\ \cline{2-10} 
 &
  15\% &
  \multicolumn{1}{c|}{0.038\%} &
  \multicolumn{1}{c|}{0.118\%} &
  \multicolumn{1}{c|}{0.018\%} &
  0.168\% &
  \multicolumn{1}{c|}{0.043\%} &
  \multicolumn{1}{c|}{0.137\%} &
  \multicolumn{1}{c|}{0.008\%} &
  0.064\% \\ \cline{2-10} 
 &
  20\% &
  \multicolumn{1}{c|}{0.038\%} &
  \multicolumn{1}{c|}{0.130\%} &
  \multicolumn{1}{c|}{0.015\%} &
  0.140\% &
  \multicolumn{1}{c|}{0.043\%} &
  \multicolumn{1}{c|}{0.141\%} &
  \multicolumn{1}{c|}{0.007\%} &
  0.065\% \\ \hlineB{2}

\multirow{4}{*}{ResNet-18-C} &
  5\% &
  \multicolumn{1}{c|}{0.029\%} &
  \multicolumn{1}{c|}{0.097\%} &
  \multicolumn{1}{c|}{0.054\%} &
  0.633\% &
  \multicolumn{1}{c|}{0.045\%} &
  \multicolumn{1}{c|}{0.153\%} &
  \multicolumn{1}{c|}{0.016\%} &
  0.137\% \\ \cline{2-10} 
 &
  \cellcolor{yellow!30}\textbf{10\%} &
  \multicolumn{1}{c|}{\cellcolor{yellow!30}\textbf{0.029\%}} &
  \multicolumn{1}{c|}{\cellcolor{yellow!30}\textbf{0.092\%}} &
  \multicolumn{1}{c|}{\cellcolor{yellow!30}\textbf{0.038\%}} &
 \cellcolor{yellow!30}\textbf{0.386\%} &
  \multicolumn{1}{c|}{\cellcolor{yellow!30}\textbf{0.045\%}} &
  \multicolumn{1}{c|}{\cellcolor{yellow!30}\textbf{0.123\%}} &
  \multicolumn{1}{c|}{\cellcolor{yellow!30}\textbf{0.011\%}} &
  \cellcolor{yellow!30}\textbf{0.076\%} \\ \cline{2-10} 
 &
  15\% &
  \multicolumn{1}{c|}{0.029\%} &
  \multicolumn{1}{c|}{0.093\%} &
  \multicolumn{1}{c|}{0.031\%} &
  0.272\% &
  \multicolumn{1}{c|}{0.045\%} &
  \multicolumn{1}{c|}{0.127\%} &
  \multicolumn{1}{c|}{0.010\%} &
  0.083\% \\ \cline{2-10} 
 &
  20\% &
  \multicolumn{1}{c|}{0.029\%} &
  \multicolumn{1}{c|}{0.071\%} &
  \multicolumn{1}{c|}{0.026\%} &
  0.300\% &
  \multicolumn{1}{c|}{0.045\%} &
  \multicolumn{1}{c|}{0.124\%} &
  \multicolumn{1}{c|}{0.008\%} &
  0.085\% \\ \hlineB{2}

\multirow{4}{*}{MobileNetV2} &
  5\% &
  \multicolumn{1}{c|}{0.031\%} &
  \multicolumn{1}{c|}{1.443\%} &
  \multicolumn{1}{c|}{0.079\%} &
  2.803\% &
  \multicolumn{1}{c|}{0.030\%} &
  \multicolumn{1}{c|}{0.893\%} &
  \multicolumn{1}{c|}{0.013\%} &
  0.983\% \\ \cline{2-10} 
 &
  \cellcolor{yellow!30}\textbf{10\%} &
  \multicolumn{1}{c|}{\cellcolor{yellow!30}\textbf{0.033\%}} &
  \multicolumn{1}{c|}{\cellcolor{yellow!30}\textbf{0.841\%}} &
  \multicolumn{1}{c|}{\cellcolor{yellow!30}\textbf{0.053\%}} &
  \cellcolor{yellow!30}\textbf{1.252\%} &
  \multicolumn{1}{c|}{\cellcolor{yellow!30}\textbf{0.030\%}} &
  \multicolumn{1}{c|}{\cellcolor{yellow!30}\textbf{0.401\%}} &
  \multicolumn{1}{c|}{\cellcolor{yellow!30}\textbf{0.009\%}} &
  \cellcolor{yellow!30}\textbf{0.372\%} \\ \cline{2-10} 
 &
  15\% &
  \multicolumn{1}{c|}{0.032\%} &
  \multicolumn{1}{c|}{0.586\%} &
  \multicolumn{1}{c|}{0.044\%} &
  0.742\% &
  \multicolumn{1}{c|}{0.030\%} &
  \multicolumn{1}{c|}{0.468\%} &
  \multicolumn{1}{c|}{0.007\%} &
  0.346\% \\ \cline{2-10} 
 &
  20\% &
  \multicolumn{1}{c|}{0.032\%} &
  \multicolumn{1}{c|}{0.465\%} &
  \multicolumn{1}{c|}{0.037\%} &
  0.720\% &
  \multicolumn{1}{c|}{0.030\%} &
  \multicolumn{1}{c|}{0.401\%} &
  \multicolumn{1}{c|}{0.006\%} &
  0.320\% \\ \hlineB{2}

\multirow{4}{*}{ResNet-18-I} &
  5\% &
  \multicolumn{1}{c|}{0.005\%} &
  \multicolumn{1}{c|}{0.071\%} &
  \multicolumn{1}{c|}{0.007\%} &
  0.168\% &
  \multicolumn{1}{c|}{0.041\%} &
  \multicolumn{1}{c|}{0.115\%} &
  \multicolumn{1}{c|}{0.016\%} &
  0.114\% \\ \cline{2-10} 
 &
  \cellcolor{yellow!30}\textbf{10\%} &
  \multicolumn{1}{c|}{\cellcolor{yellow!30}\textbf{0.005\%}} &
  \multicolumn{1}{c|}{\cellcolor{yellow!30}\textbf{0.070\%}} &
  \multicolumn{1}{c|}{\cellcolor{yellow!30}\textbf{0.005\%}} &
  \cellcolor{yellow!30}\textbf{0.212\%} &
  \multicolumn{1}{c|}{\cellcolor{yellow!30}\textbf{0.041\%}} &
  \multicolumn{1}{c|}{\cellcolor{yellow!30}\textbf{0.096\%}} &
  \multicolumn{1}{c|}{\cellcolor{yellow!30}\textbf{0.010\%}} &
  \cellcolor{yellow!30}\textbf{0.061\%} \\ \cline{2-10} 
 &
  15\% &
  \multicolumn{1}{c|}{0.005\%} &
  \multicolumn{1}{c|}{0.054\%} &
  \multicolumn{1}{c|}{0.004\%} &
  0.108\% &
  \multicolumn{1}{c|}{0.041\%} &
  \multicolumn{1}{c|}{0.083\%} &
  \multicolumn{1}{c|}{0.008\%} &
  0.062\% \\ \cline{2-10} 
 &
  20\% &
  \multicolumn{1}{c|}{0.005\%} &
  \multicolumn{1}{c|}{0.045\%} &
  \multicolumn{1}{c|}{0.003\%} &
  0.082\% &
  \multicolumn{1}{c|}{0.042\%} &
  \multicolumn{1}{c|}{0.089\%} &
  \multicolumn{1}{c|}{0.006\%} &
  0.032\% \\ \hlineB{2}

\multirow{4}{*}{ResNet-34} &
  5\% &
  \multicolumn{1}{c|}{0.003\%} &
  \multicolumn{1}{c|}{0.061\%} &
  \multicolumn{1}{c|}{0.005\%} &
  0.140\% &
  \multicolumn{1}{c|}{0.045\%} &
  \multicolumn{1}{c|}{0.136\%} &
  \multicolumn{1}{c|}{0.016\%} &
  0.087\% \\ \cline{2-10} 
 &
  \cellcolor{yellow!30}\textbf{10\%} &
  \multicolumn{1}{c|}{\cellcolor{yellow!30}\textbf{0.003\%}} &
  \multicolumn{1}{c|}{\cellcolor{yellow!30}\textbf{0.059\%}} &
  \multicolumn{1}{c|}{\cellcolor{yellow!30}\textbf{0.003\%}} &
  \cellcolor{yellow!30}\textbf{0.084\%} &
  \multicolumn{1}{c|}{\cellcolor{yellow!30}\textbf{0.045\%}} &
  \multicolumn{1}{c|}{\cellcolor{yellow!30}\textbf{0.127\%}} &
  \multicolumn{1}{c|}{\cellcolor{yellow!30}\textbf{0.011\%}} &
  \cellcolor{yellow!30}\textbf{0.097\%} \\ \cline{2-10} 
 &
  15\% &
  \multicolumn{1}{c|}{0.003\%} &
  \multicolumn{1}{c|}{0.054\%} &
  \multicolumn{1}{c|}{0.002\%} &
  0.089\% &
  \multicolumn{1}{c|}{0.046\%} &
  \multicolumn{1}{c|}{0.122\%} &
  \multicolumn{1}{c|}{0.009\%} &
  0.080\% \\ \cline{2-10} 
 &
  20\% &
  \multicolumn{1}{c|}{0.003\%} &
  \multicolumn{1}{c|}{0.048\%} &
  \multicolumn{1}{c|}{0.002\%} &
  0.063\% &
  \multicolumn{1}{c|}{0.045\%} &
  \multicolumn{1}{c|}{0.120\%} &
  \multicolumn{1}{c|}{0.007\%} &
  0.061\% \\ \hline
  
\end{tabular}%
}
\end{table*}

This subsection presents the results for sampling DeepVigor+ and compares its VF results against the complete analysis to show how accurate sampling DeepVigor+ is. Channel Vulnerability Factor (CVF) and Layer Vulnerability Factor (LVF) are derived as described in subsection \ref{subsec:deepvigor+} and the maximum and mean absolute error for obtained CVFs and LVFs in complete and sampling DeepVigor+ for neurons and filters are presented separately. In sampling DeepVigor+, \textit{channel sampling ratio} is explored. 

Table \ref{tab:sampling-error} indicates the absolute error results over various \textit{channel sampling ratio}s for CNNs under study, in both neurons and filters analysis. Each sampling analysis is repeated $50$ times to observe the effect of random selections in stratified sampling. Based on the results, the minimum absolute error throughout the experiments for both CVF and LVF is very close to $0$. As observed, the difference between obtained CVFs and LVFs throughout the results is minimal, demonstrating the effectiveness of the exploited stratified sampling. 

In all experiments, the MAE for CVF does not vary since the sampling method within channels is similar (i.e., the logarithm of the number of OFMaps). Also, it is observed that the MAE CVF is less than $0.065\%$ throughout the experiments for both activations and filters sampling analysis. It means that the averaged VF for the logarithm-based random sampling within each channel results in a highly accurate CVF. This phenomenon is a result of the unified distribution of weights within a channel in a pre-trained CNN. The maximum observed error in neuron analysis is $0.004\%$ to $1.443\%$ throughout CNNs, whereas the mean error remains low, meaning that for most of the channels, obtained CVFs are highly accurate, leading to an overall high accuracy for CVFs. 

On the other hand, \textit{channel sampling ratio} directly affects the accuracy of LVF calculations. As observed in Table \ref{tab:sampling-error}, LVF error decreases with the increase of \textit{channel sampling ratio}. Considering both MAE and maximum error for LVF, a $10\%$ \textit{channel sampling ratio} can guarantee a minimal error for VF calculations. 
Based on the MAE of complete DeepVigor+ compared to exhaustive FI Table \ref{tab:FI-validation}, sampling DeepVigor+ with a $10\%$ channel sampling ratio ensures that all the overall errors of the obtained VFs for all CNNs will be lower than $1\%$.

In conclusion, the proposed stratified sampling in DeepVigor+ results in highly accurate VF for channels and layers obtained from both activations and filter analysis with a \textit{channel sampling ratio} of $10\%$. Sampling DeepVigor+ results in a higher error than state-of-the-art statistical FI approaches, such as data-aware and data-unaware, yet it meets the above-mentioned requirement of average error for the resilience study, which is $1\%$.

\subsection{Run-Time and Scalability Investigation}

To show the effectiveness of DeepVigor+ in terms of complexity, scalability, and execution time against DeepVigor and FI, first, we investigate the complexity of each method based on the required number of simulations (i.e., forward pass executions) to obtain VF. Then we present the execution time for the complete and sampling DeepVigor+ on an NVIDIA A100 GPU.

Exhaustive FI is the most accurate method for determining precise VFs. In exhaustive FI, the required number of simulations equals the number of activations/weights times bit-width (i.e., 32 bits). Therefore, its complexity is proportional linearly to the size of CNNs. On the other hand, DeepVigor {\cite{ahmadilivani2023deepvigor}} is faster than exhaustive FI to obtain the VF for activations. To obtain a vulnerability value range, DeepVigor requires $9.1$ simulations per neuron, on average.

Whereas the complete DeepVigor+ analysis estimates VFs with high accuracy and significantly lower complexity. Although its complexity is affected by the size of CNNs, DeepVigor+ analysis exploits various optimizations to reduce the number of simulations, resulting in significantly lower complexity than exhaustive FI and DeepVigor. This is evidenced by the results in Table \ref{tab:exFI-compdv} where the required number of simulations is compared between Exhaustive FI, DeepVigor, and complete DeepVigor+ analysis, for activations and filters separately.

\begin{table*}[h!]
\caption{Number of simulations for exhaustive FI, DeepVigor, and complete DeepVigor+ for activations and filters analysis.}
%\label{tab:my-table}
%\resizebox{\columnwidth}{!}{%
\centering
\begin{tabular}{|c|ccc||cc|}
\hline
             & \multicolumn{3}{c||}{Activations}      & \multicolumn{2}{c|}{Filters} \\ \hline 
CNNs         & \multicolumn{1}{c|}{Exhaustive FI} & \multicolumn{1}{c|}{DeepVigor}  & Complete DeepVigor+   & \multicolumn{1}{c|}{Exhaustive FI}    &   Complete DeepVigor+ \\ \hline \hline
VGG-11       & \multicolumn{1}{c|}{7,438,336}     & \multicolumn{1}{c|}{5,124,816}  &  \textbf{1,636,414}   &  \multicolumn{1}{c|}{294,967,296}   &   \textbf{11,035}  \\ \hline
VGG-16       & \multicolumn{1}{c|}{5,931,008}     & \multicolumn{1}{c|}{3,586,312}  &  \textbf{2,018,447}   &  \multicolumn{1}{c|}{470,734,848}   &   \textbf{13,704}  \\ \hline
ResNet-18-C  & \multicolumn{1}{c|}{21,331,968}    & \multicolumn{1}{c|}{9,941,732}  &  \textbf{3,387,189}   &  \multicolumn{1}{c|}{357,095,424}   &   \textbf{15,906}  \\ \hline
MobileNet-V2 & \multicolumn{1}{c|}{27,330,048}    & \multicolumn{1}{c|}{12,958,887} &  \textbf{3,029,125}   &  \multicolumn{1}{c|}{74,176,512}    &   \textbf{42,234}  \\ \hline
ResNet-18-I  & \multicolumn{1}{c|}{69,844,992}    & \multicolumn{1}{c|}{45,083,325} &  \textbf{20,730,314}  &  \multicolumn{1}{c|}{357,341,184}   &   \textbf{23,450}  \\ \hline
ResNet-34    & \multicolumn{1}{c|}{109,985,792}   & \multicolumn{1}{c|}{66,303,692} &  \textbf{28,821,489}  & \multicolumn{1}{c|}{680,564,736}    &  \textbf{43,098}  \\ \hline
\end{tabular}%
%}
\label{tab:exFI-compdv}
\end{table*}

As observed, for each CNN, DeepVigor+ complete analysis requires significantly fewer simulations than exhaustive FI and DeepVigor, in both activations and filters analysis. Throughout the results, activation analysis by complete DeepVigor+ is $2.93$ to $9.02$ times faster than exhaustive FI. Furthermore, DeepVigor+ outperforms DeepVigor in activations analysis, requiring $1.77$ to $4.28$ times fewer simulations. Note that the accuracy of DeepVigor to obtain VF is $<0.1\%$ while the accuracy of DeepVigor+ is $<1\%$.  

According to our detailed investigations, activation analysis by DeepVigor+ requires $6$ forward simulations per neuron, on average, throughout the CNNs under study. In this regard, the introduced loss function to obtain the vulnerability values (box 4 in Fig. \ref{fig:DV+flowchart}) contributes to skipping the analysis for up to $26\%$ of neurons. Moreover, the vulnerability value can be obtained by a single forward simulation for up to $55\%$ of neurons due to separating VVSS exploration (boxes 7 and 8 in Fig. \ref{fig:DV+flowchart}).

Moreover, in the complete filters analysis, the DeepVigor+ fault propagation modeling implies an even more significant impact on the number of simulations in the orders of magnitude. DeepVigor+ can derive VF for filters $1,756$ to $34,350$ times faster than exhaustive FI. To obtain the vulnerability values for channels, up to $1\%$ of channels are skipped by leveraging the loss function, and up to $34\%$ of channels require one forward simulation. These results indicate that complete DeepVigor+ provides accurate VF for neurons and channels of CNNs with significantly lower complexity and shorter execution time than exhaustive FI enabled by its optimal fault propagation modeling and analysis.

\begin{table*}[t]
\caption{Comparison of required simulations for different statistical FI methods based on {\cite{ruospo2023assessing}} and sampling DeepVigor+.}
\label{tab:simulation-count}
%\resizebox{\textwidth}{!}{%
\centering
\begin{tabular}{|c|cccc||cccc|}
\hline
 &
  \multicolumn{4}{c||}{Activations Analysis} &
  \multicolumn{4}{c|}{Filters Analysis} \\ \hline
\begin{tabular}[c]{@{}c@{}}Analysis \\ method\end{tabular} &
  \multicolumn{1}{c|}{Layer-wise} &
  \multicolumn{1}{c|}{Data-unaware} &
  \multicolumn{1}{c|}{Data-aware} &
  \begin{tabular}[c]{@{}c@{}} Sampling \\ DeepVigor+ \\ CSR = 10\%\end{tabular} &
  \multicolumn{1}{c|}{Layer-wise} &
  \multicolumn{1}{c|}{Data-unaware} &
  \multicolumn{1}{c|}{Data-aware} &
  \begin{tabular}[c]{@{}c@{}} Sampling \\DeepVigor+ \\ CSR = 10\% \end{tabular} \\ \hline \hline

VGG-11 &
  \multicolumn{1}{c|}{74,863} &
  \multicolumn{1}{c|}{1,562,657} &
  \multicolumn{1}{c|}{71,173} &
  \textbf{4,596} &
  \multicolumn{1}{c|}{75,351} &
  \multicolumn{1}{c|}{2,141,913} &
  \multicolumn{1}{c|}{66,934} &
  \textbf{1,038} \\ \hline

VGG-16 &
  \multicolumn{1}{c|}{117,992} &
  \multicolumn{1}{c|}{1,839,889} &
  \multicolumn{1}{c|}{106,513} &
  \textbf{10,283} &
  \multicolumn{1}{c|}{123,266} &
  \multicolumn{1}{c|}{3,588,834} &
  \multicolumn{1}{c|}{112,151} &
  \textbf{1,476} \\ \hline

ResNet-18-C &
  \multicolumn{1}{c|}{188,644} &
  \multicolumn{1}{c|}{4,264,406} &
  \multicolumn{1}{c|}{181,911} &
  \textbf{15,996} &
  \multicolumn{1}{c|}{189,772} &
  \multicolumn{1}{c|}{5,253,096} &
  \multicolumn{1}{c|}{164,159} &
  \textbf{1,706} \\ \hline

MobileNetV2 &
  \multicolumn{1}{c|}{493,871} &
  \multicolumn{1}{c|}{8,402,977} &
  \multicolumn{1}{c|}{452,650} &
  \textbf{22,077} &
  \multicolumn{1}{c|}{475,407} &
  \multicolumn{1}{c|}{8,307,671} &
  \multicolumn{1}{c|}{259,614} &
  \textbf{4,364} \\ \hline

ResNet-18-I &
  \multicolumn{1}{c|}{191,205} &
  \multicolumn{1}{c|}{5,407,659} &
  \multicolumn{1}{c|}{189,325} &
  \textbf{21,680} &
  \multicolumn{1}{c|}{190,896} &
  \multicolumn{1}{c|}{5,358,315} &
  \multicolumn{1}{c|}{167,447} &
  \textbf{2,178} \\ \hline

ResNet-34 &
  \multicolumn{1}{c|}{344,042} &
  \multicolumn{1}{c|}{9,624,374} &
  \multicolumn{1}{c|}{340,383} &
  \textbf{39,002} &
  \multicolumn{1}{c|}{344,285} &
  \multicolumn{1}{c|}{10,031,494} &
  \multicolumn{1}{c|}{313,484} &
  \textbf{4,003} \\ \hline
\end{tabular}%
%}
\label{tab:statFI-sampdv}
\end{table*}

On the other hand, sampling DeepVigor+ is proposed to further reduce the execution time and complexity of resilience analysis and achieve a scalable method. To show its performance against Statistical FI (SFI), Table \ref{tab:statFI-sampdv} compares the number of simulations for various state-of-the-art SFI methods proposed in \cite{ruospo2023assessing} with sampling DeepVigor+. As presented, data-aware SFI leads to the least number of executions in the FI-based simulation. It is observed that DeepVigor+ sampling activations analysis with $10\%$ channel sampling ratio leads to $8.72$ to $20.5$ times fewer simulations compared to data-aware SFI. For the filters analysis, DeepVigor+ obtains their VF with $59.4$ up to $96.2$ times fewer simulations than data-aware SFI, which is currently accepted to be the fastest state-of-the-art approach. 

To obtain the Model Vulnerability Factor (MVF) both activations and filters should be analyzed separately, based on Eq. \eqref{eq:mvf-total}. Therefore, sampling DeepVigor+ accelerates the process, ranging from $14.9$ up to $26.9$ times throughout the CNNs. The scalability and speed of the method are achieved by both channel sampling ratio and logarithmic sampling within them. It can be observed that with the remarkable growth of CNNs under analysis in their number of parameters, the number of simulations in DeepVigor+ does not grow linearly.

To demonstrate the execution time of DeepVigor+ analysis, we employed an A100 NVIDIA GPU and performed sampling DeepVigor+ for activations and filters with different channel sampling ratios. Table \ref{tab:time-act} and Table \ref{tab:time-weight} present the average execution time over $50$ executions of the method for complete and sampling DeepVigor+ with different channel sampling ratios, for activations and filters, respectively. It is observed that VFs can be obtained in a few minutes for any CNN. 

Considering $10\%$ channel sampling ratio for both activations and filters analysis, the total MVF for VGG-11, VGG-16, ResNet-18-C, MobileNet-V2, ResNet-18-I and ResNet-34 is obtained in almost $8.7$ minutes, $9.6$ minutes, $12.7$ minutes, $43.5$ minutes, $19.4$ minutes, and $46$ minutes, respectively. It is worth mentioning that the complete DeepVigor+ analysis for the CNNs under study takes almost $22$ hours for VGG-16 and $18.5$ days for ResNet-34, on the same GPU. 
Fast execution and accurate estimation of VF obtained by sampling DeepVigor+ analysis provide a remarkable opportunity for a high-speed resilience analysis for any known CNN.

\begin{table}[h!]
\small
\centering
\caption{Average execution time on A100 GPU for DeepVigor+ activations analysis, with different Channel Sampling Ratios (CSR).}
\resizebox{\columnwidth}{!}{%
\begin{tabular}{cccccc}\toprule\toprule
CNN      &   CSR = 5\%   &   \underline{ CSR = 10\%}   &   CSR = 15\%  &   CSR = 20\%   & \begin{tabular}[c]{@{}c@{}} Neurons \\ complete \\ analysis \end{tabular}   \\ \toprule\toprule
VGG-11      &   \begin{tabular}[c]{@{}c@{}}  203 sec \\ ($\approx$ 3.4 min) \end{tabular} &  \begin{tabular}[c]{@{}c@{}}  415 sec \\ \cellcolor{yellow!30}($\approx$ \textbf{6.9 min}) \end{tabular}  &  \begin{tabular}[c]{@{}c@{}}   625 sec \\ ($\approx$ 10.4 min) \end{tabular}   &   \begin{tabular}[c]{@{}c@{}}   838 sec \\ ($\approx$ 13.9 min) \end{tabular}   &  \begin{tabular}[c]{@{}c@{}}  66,083 sec \\ ($\approx$ 0.7 days) \end{tabular} 
   \\ \toprule
   
VGG-16      &   \begin{tabular}[c]{@{}c@{}}  223 sec \\ ($\approx$ 3.7 min) \end{tabular}   &  \begin{tabular}[c]{@{}c@{}}  453 sec \\ \cellcolor{yellow!30}($\approx$ \textbf{7.5 min}) \end{tabular}   &  \begin{tabular}[c]{@{}c@{}}  676 sec \\ ($\approx$ 11 min) \end{tabular}   &   \begin{tabular}[c]{@{}c@{}}  915 sec \\ ($\approx$ 15 min) \end{tabular}     &  \begin{tabular}[c]{@{}c@{}}  43,326 sec \\ ($\approx$ 0.5 days) \end{tabular}  \\ \toprule

ResNet-18-C &   \begin{tabular}[c]{@{}c@{}}  284 sec \\ ($\approx$ 4.7 min) \end{tabular}   &  \begin{tabular}[c]{@{}c@{}}  580 sec \\ \cellcolor{yellow!30}($\approx$ \textbf{9.5 min}) \end{tabular}   &  \begin{tabular}[c]{@{}c@{}}  876 sec \\ ($\approx$ 14.5 min) \end{tabular}    &   \begin{tabular}[c]{@{}c@{}}  1,171 sec \\ ($\approx$ 19.5 min) \end{tabular}    &  \begin{tabular}[c]{@{}c@{}}  94,939 \\ ($\approx$ 1.1 days) \end{tabular} 
   \\ \toprule
   
MobileNetV2 &   \begin{tabular}[c]{@{}c@{}}  1,037 sec \\ ($\approx$ 17 min) \end{tabular}   &  \begin{tabular}[c]{@{}c@{}}  2,097 sec \\ \cellcolor{yellow!30}($\approx$ \textbf{35 min}) \end{tabular}  &  \begin{tabular}[c]{@{}c@{}}  3,153 sec \\ ($\approx$ 52.5 min) \end{tabular}   &   \begin{tabular}[c]{@{}c@{}}  4,071 sec \\ ($\approx$ 68 min) \end{tabular}    &  \begin{tabular}[c]{@{}c@{}}  211,868 \\ ($\approx$ 2.4 days) \end{tabular}  \\ \toprule

ResNet-18-I &   \begin{tabular}[c]{@{}c@{}}  448 sec \\ ($\approx$ 7.4 min) \end{tabular}   &  \begin{tabular}[c]{@{}c@{}}  917 sec \\ \cellcolor{yellow!30}($\approx$ \textbf{15 min}) \end{tabular}   &  \begin{tabular}[c]{@{}c@{}}  1,390 sec \\ ($\approx$ 23 min) \end{tabular}   &   \begin{tabular}[c]{@{}c@{}}  1,866 sec \\ ($\approx$ 31 min) \end{tabular}     &  \begin{tabular}[c]{@{}c@{}}  634,872 \\ ($\approx$ 7.3 days) \end{tabular} \\ \toprule

ResNet-34   &  \begin{tabular}[c]{@{}c@{}}  1,100 sec \\ ($\approx$ 18 min) \end{tabular}   &  \begin{tabular}[c]{@{}c@{}}  2,237 sec \\ \cellcolor{yellow!30}($\approx$ \textbf{37 min}) \end{tabular}   &  \begin{tabular}[c]{@{}c@{}} 3398 sec \\ ($\approx$ 56 min) \end{tabular}    &   \begin{tabular}[c]{@{}c@{}} 4549  sec \\ ($\approx$ 75 min) \end{tabular}    &  \begin{tabular}[c]{@{}c@{}} 1,402,362 \\ ($\approx$ 16.2 days) \end{tabular}  \\ \bottomrule
\end{tabular}
}
\label{tab:time-act}
\end{table}

\begin{table}[h!]
\small
\centering
\caption{Average execution time on A100 GPU for DeepVigor+ filters analysis, with different Channel Sampling Ratios (CSR).}
\resizebox{\columnwidth}{!}{%
\begin{tabular}{cccccc}\toprule\toprule
CNN      &    CSR = 5\%   &    \underline{CSR = 10\%}    &    CSR = 15\%    &    CSR = 20\%    & \begin{tabular}[c]{@{}c@{}} Filters \\ complete \\ analysis \end{tabular}   \\ \toprule\toprule
VGG-11       &   \begin{tabular}[c]{@{}c@{}}  57 sec \\ ($\approx$ 0.95 min) \end{tabular}   &  \begin{tabular}[c]{@{}c@{}}  111 sec \\ \cellcolor{yellow!30}($\approx$ \textbf{1.8 min}) \end{tabular}  &  \begin{tabular}[c]{@{}c@{}}  170 sec \\ ($\approx$ 2.8 min) \end{tabular}   &   \begin{tabular}[c]{@{}c@{}}  219 sec \\ ($\approx$ 3.6 min) \end{tabular}   &  \begin{tabular}[c]{@{}c@{}}  23,031 sec \\ ($\approx$ 0.26 days) \end{tabular}    \\ \toprule
VGG-16       &   \begin{tabular}[c]{@{}c@{}}  61 sec \\ ($\approx$ 1 min) \end{tabular}   &  \begin{tabular}[c]{@{}c@{}}  125 sec \\ \cellcolor{yellow!30}($\approx$ \textbf{2.1 min}) \end{tabular}   &  \begin{tabular}[c]{@{}c@{}}  168 sec \\ ($\approx$ 3.1 min) \end{tabular}    &   \begin{tabular}[c]{@{}c@{}}  245 sec \\ ($\approx$ 4.1 min) \end{tabular}    &  \begin{tabular}[c]{@{}c@{}}  35,634 sec \\ ($\approx$ 0.4 days) \end{tabular}   \\ \toprule
ResNet-18-C  &   \begin{tabular}[c]{@{}c@{}}  92 sec \\ ($\approx$ 1.5 min) \end{tabular}   &  \begin{tabular}[c]{@{}c@{}}  184 sec \\ \cellcolor{yellow!30}($\approx$ \textbf{3 min}) \end{tabular}  &  \begin{tabular}[c]{@{}c@{}}  275 sec \\ ($\approx$ 4.6 min) \end{tabular}    &   \begin{tabular}[c]{@{}c@{}}  367 sec \\ ($\approx$ 6.1 min) \end{tabular}    &  \begin{tabular}[c]{@{}c@{}}  60,410 sec \\ ($\approx$ 0.7 days) \end{tabular}   \\ \toprule
MobileNetV2  &  \begin{tabular}[c]{@{}c@{}}  257 sec \\ ($\approx$ 4.3 min) \end{tabular}   &  \begin{tabular}[c]{@{}c@{}}  513 sec \\ \cellcolor{yellow!30}($\approx$ \textbf{8.5 min}) \end{tabular}   &  \begin{tabular}[c]{@{}c@{}}  759 sec \\ ($\approx$ 12.6 min) \end{tabular}    &   \begin{tabular}[c]{@{}c@{}}  1,025 sec \\ ($\approx$ 17.1 min) \end{tabular}   &  \begin{tabular}[c]{@{}c@{}}  39,713 sec \\ ($\approx$ 0.46 days) \end{tabular}   \\ \toprule
ResNet-18-I  &  \begin{tabular}[c]{@{}c@{}} 117 sec \\ ($\approx$ 1.9 min) \end{tabular}   & \begin{tabular}[c]{@{}c@{}} 250 sec \\ \cellcolor{yellow!30}($\approx$ \textbf{4.1 min}) \end{tabular}    &  \begin{tabular}[c]{@{}c@{}} 370 sec \\ ($\approx$ 6.1 min) \end{tabular}    &   \begin{tabular}[c]{@{}c@{}} 506 sec \\ ($\approx$ 8.4 min) \end{tabular}    &   \begin{tabular}[c]{@{}c@{}}  134,164 sec \\ ($\approx$ 1.5 days) \end{tabular}  \\ \toprule
ResNet-34    &  \begin{tabular}[c]{@{}c@{}} 259 sec \\ ($\approx$ 4.3 min) \end{tabular}   &  \begin{tabular}[c]{@{}c@{}} 524 sec \\ \cellcolor{yellow!30}($\approx$ \textbf{8.7 min}) \end{tabular}   &  \begin{tabular}[c]{@{}c@{}} 795 sec \\ ($\approx$ 13.2 min) \end{tabular}    &   \begin{tabular}[c]{@{}c@{}} 1,064 sec \\ ($\approx$ 17.7 min) \end{tabular}   &  \begin{tabular}[c]{@{}c@{}}  201,305 sec \\ ($\approx$ 2.3 days) \end{tabular}  \\ \bottomrule
\end{tabular}
}
\label{tab:time-weight}
\end{table}

Furthermore, the implementation of DeepVigor+ is resource-efficient and does not impose significant GPU memory demands. Across all experiments, using a $10\%$ \textit{channel sampling ratio} and a \textit{batch-size} of $100$, the GPU memory consumption ranges from a minimum of $4$ GB to a maximum of $5$ GB. This demonstrates both the computational efficiency and scalability of the proposed method. As a result, DeepVigor+ can be readily deployed on a wide range of hardware platforms, including GPUs with limited memory capacity, making it accessible for broader adoption in both academic and industrial settings.

\subsection{Vulnerability Visualization and Comparison for CNNs}

\begin{figure}[h!]
\vspace{-2mm}
    \includegraphics[width=0.48\textwidth]{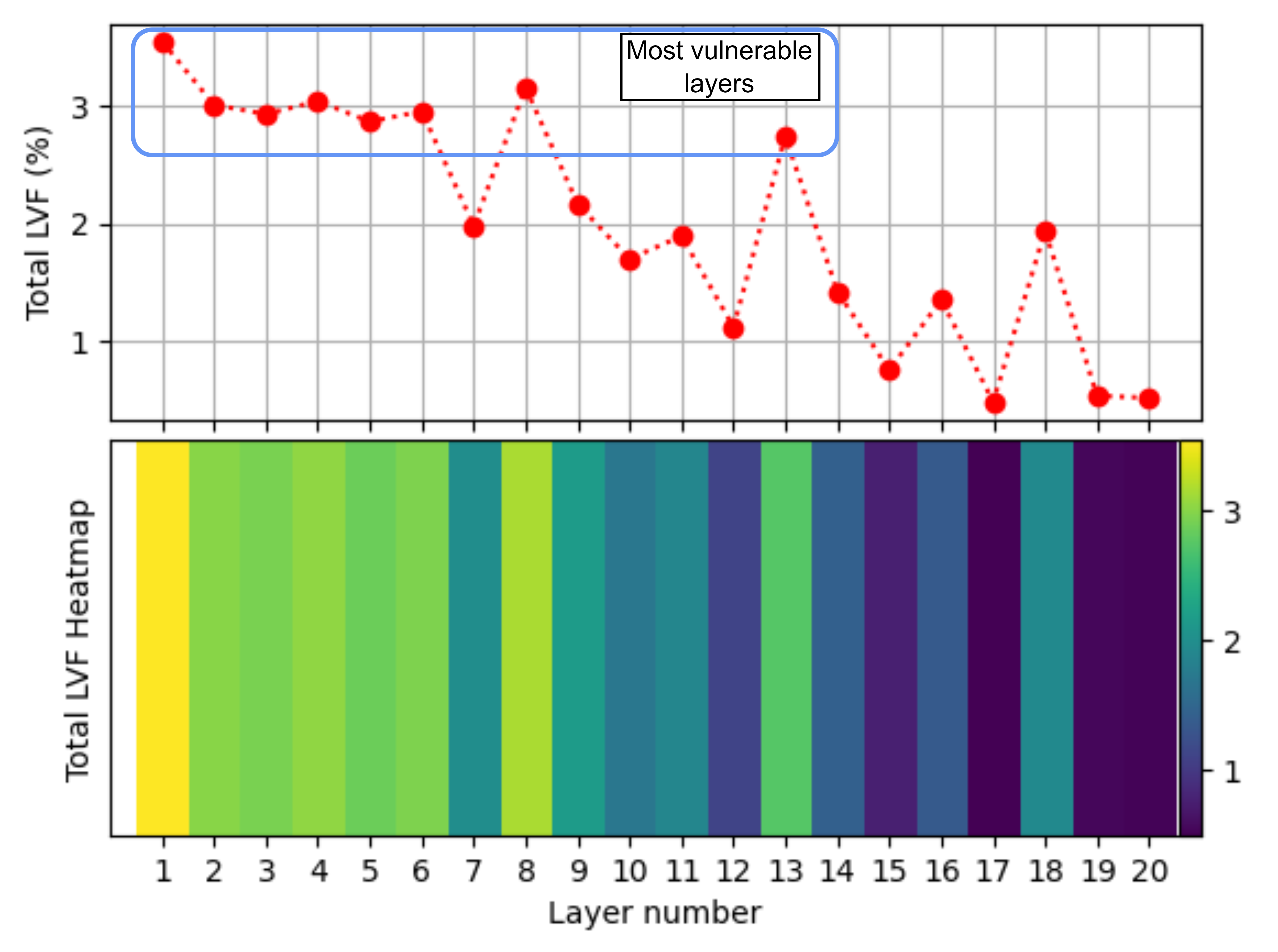}
\centering
\caption {LVF visualization and comparison for ResNet-18 on ImageNet. \vspace{-5mm}}
\label{fig:resnet-lvf}
\end{figure}

It has been demonstrated that VF for CNN layers and the entire model can be obtained accurately in a few minutes by DeepVigor+. The obtained VF results by DeepVigor+ can be used to visualize the vulnerability of layers within a CNN and identify the most vulnerable ones, leading to efficient protection mechanisms. Fig. \ref{fig:resnet-lvf} illustrates the comparison of total LVF for ResNet-18 trained on ImageNet, while total LVF for each layer is obtained based on the numerator of Eq. \eqref{eq:mvf-total}, which considers the LVF of both weights and activations. This visualization sketches how much each layer is vulnerable to single faults compared to the others. As observed, the first layers are generally more vulnerable than the latter ones. Layers 1 to 6, 8, and 13 demonstrate more vulnerability, and they need stronger fault mitigation techniques, while in other layers, the costs of implementing fault tolerance can be saved.  
%Therefore, DeepVigor+ enables vulnerability visualization and comparison within a CNN. 

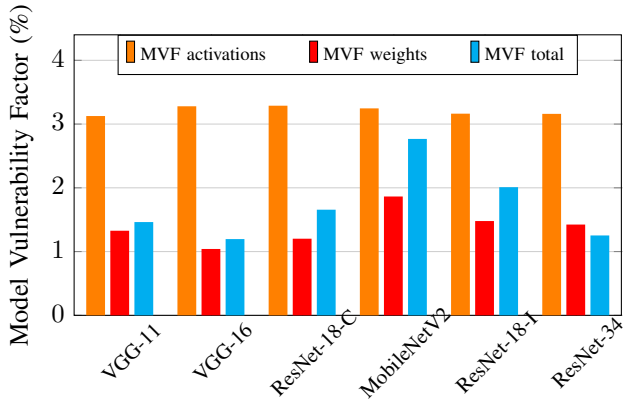
\begin{figure}[t!]
    \centering
    \begin{tikzpicture}
  \centering
  \begin{axis}[
        width=\columnwidth,
        height=0.6\columnwidth,
        ybar,
        bar width=0.25cm,
        major grid style={draw=white},
        grid=minor,
        minor ytick={0, 1, 2, 3, 4},
        grid style={line width=.1pt, draw=gray!40},
        major grid style={line width=.2pt,draw=gray!50},
        enlarge y limits={value=.1,upper},
        ymin=0, ymax=4,
        xmin=1, xmax=6,
        tickwidth=0pt,
        enlarge x limits=true,
        legend style={
            at={(0.5,-0.2)},
            anchor=north,
            legend columns=-1,
            /tikz/every even column/.append style={column sep=0.5cm}
        },
       ylabel={Model Vulnerability Factor (\%)},
       legend style={at={(0.5,1)},
            anchor=north,legend columns=-1, font=\fontsize{7}{4}\selectfont},
        label style={font=\small},
       xtick={1, 2, 3, 4, 5, 6}, 
       ytick={0, 1, 2, 3, 4},
       x tick label style={rotate=45,anchor=north,font=\footnotesize, xshift=-1.5ex , yshift = -1.5ex},
        xticklabels = {VGG-11, VGG-16, ResNet-18-C, MobileNetV2, ResNet-18-I, ResNet-34},
        yticklabels = {\strut $0$, \strut $1$,\strut $2$, \strut $3$, \strut $4$},
    ]
    \addplot [draw=none, fill=orange] coordinates {(1, 3.126) (2, 3.279) (3, 3.288) (4, 3.246) (5, 3.163) (6, 3.16)};
    \addplot [draw=none, fill=red] coordinates {(1, 1.327) (2, 1.041) (3, 1.202) (4, 1.864) (5, 1.479) (6, 1.253)};
    \addplot [draw=none, fill=cyan] coordinates {(1, 1.463) (2, 1.196) (3, 1.657) (4, 2.767) (5, 2.010) (6, 1.424)};
   \legend{MVF activations, MVF weights, MVF total} 
  \end{axis}
  \end{tikzpicture}
    \caption{MVF comparison for CNNs based on activations, filters, and the entire model derived by DeepVigor+. \vspace{-4mm}}
    \label{fig:mvf-cmp}
\end{figure}

Furthermore, DeepVigor+ results in total MVF based on Eq. \eqref{eq:mvf-total}, which is the weighted average of obtained LVFs, providing a comprehensive examination of vulnerability between different CNNs. Fig. \ref{fig:mvf-cmp} indicates MVFs for activations and filters of CNNs, separately, as well as their total MVF. As a result, it is observed that activations are more vulnerable than weights. However, weights generally contribute more to the total MVF since their number is higher than that of activations in CNNs. Except in MobilNetV2, the number of weights is smaller, so the activations have a more significant impact on the total MVF. Based on total MVF, VGG-16 is the least vulnerable CNN (MVF = $1.19\%$), and MobileNetV2 is the most vulnerable one (MVF = $2.76\%$). 

This visualization also enables analyzing the impact of architectural characteristics of CNNs on their fault resilience. As observed, in both VGG and ResNet, their deeper versions, VGG-16 and ResNet-34, possess a smaller MVF than their shallower counterparts, VGG-11 and ResNet-18-I, respectively. Also, the specific architecture of MobileNet-V2, which contains inverted residual blocks and depthwise separable convolutions, leads to more vulnerability than other CNN architectures.

\subsection{Impact of Input Data on the Quality of Results}

\begin{figure}[t!]
\captionsetup{justification=centering}
\centering

\begin{tikzpicture}
    \begin{axis}[
     width=0.9\columnwidth,
         height=0.7\columnwidth,
         font=\footnotesize,
        ybar,
        bar width=7pt,
        % (changed from `xtick`)
        scaled x ticks = false,
        xtick={1, 2, 3, 4, 5, 6},
        xticklabels = {VGG-11, VGG-16, ResNet-18-C, MobileNet-V2, ResNet-18-I, ResNet-34},
        xlabel near ticks,
        label style={font=\small},
       x tick label style={rotate=45,anchor=north,font=\footnotesize, xshift=-2ex , yshift = -1.5ex},
        %xtick distance=1,
        %xlabel=CNNs,
        ylabel=Total MVF (\%),
        enlarge x limits={abs=0.5},
        ymin=0,
        scaled ticks=false,
        % remove the `xticks`
        xtick style={
            /pgfplots/major tick length=0pt,
        },
    ]

        \addplot+ [
            error bars/.cd,
                y dir=both, y explicit,
        ] table [
            y error plus=ey+,
            y error minus=ey-,
        ] {
            x       y           ey+             ey-
            1       1.46558     0.01397358      0.0090417
            2       1.20646     0.014920        0.02394
            3       1.64874     0.023586        0.02821138
            4       2.798617    0.04433         0.0309016
            5       2.020028    0.01624         0.009748
            6       1.258157    0.01706         0.0075
        };  
    \end{axis}
\end{tikzpicture}
\caption{Total MVF variation over different batches of data for all CNNs, when batch-size = $100$.}
\vspace{-3mm}
\label{fig:batch-data}

\end{figure}
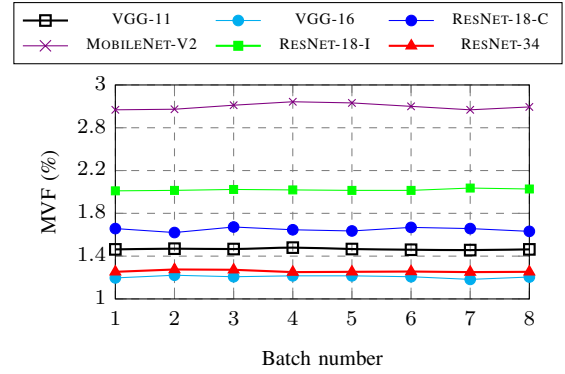

To obtain the VFs in DeepVigor+, this paper has considered one batch of $100$ input data. To investigate the impact of data on the quality of analysis results, we repeat the experiments using different batches of data with $100$ input data and derive the total MVF of the CNNs. Fig. \ref{fig:batch-data} illustrates the obtained total MVF for CNNs over $8$ different batches of data. As observed, the variation between the total MVF for each CNN is negligible across different batches of data, demonstrating that analyzing CNNs' resilience with a single batch of data provides confident results.

%\hl{Furthermore, these results demonstrate that 100 input data can be an effective representation of a dataset for such analysis.}  

To evaluate the sensitivity of DeepVigor+ to the number of input samples, we conducted experiments using varying batch sizes ranging from $20$ to $200$ across all CNN models. The results revealed a negligible variation in the computed MVF, with differences consistently below $0.05\%$ across all configurations. These findings indicate that the resilience analysis performed by DeepVigor+ is largely insensitive to the batch size and the specific set of input images used, demonstrating the robustness and stability of the method.

%% file: sections/6-conclusion.tex
\section{Discussion} \label{sec:discussion}

\subsection{Constraints and Considerations}

As shown, DeepVigor+ achieves a fast, scalable, and accurate resilience analysis for CNNs. Yet, there are some constraints in this method to be considered and extended in future research:

\begin{itemize}
    \item It is assumed that the parameters within the layers of CNNs under analysis are uniformly distributed among channels and their positions are not rearranged after training. In opposite cases, a higher channel sampling ratio is needed to obtain accurate VF results. 
    
    \item DeepVigor+ supports the single-bit fault model in input activations and weights of convolutional layers, and the obtained VFs correspond to this fault model. For multi-bit fault models, the corresponding error propagation should be applied. 
    
    \item The error propagation analysis presented in this paper is based on 32-bit floating-point data representation. However, the same concept can be extended and applied to fixed-point and integer data representations for QNNs resilience analysis, as in \cite{ahmadilivani2023enhancing}.

    \item DeepVigor+ is presented for CNNs on the image classification application. Yet, the presented theories and techniques for fault propagation are valid for other types of DL models and applications, such as RNNs and Transformers in object detection, language processing, etc. To apply DeepVigor+ to other DL models and applications, the error propagation mechanism and failure analysis should be extended based on their unique operations. 
    
\end{itemize}

\subsection{Applications, Impacts, and Future Directions}

The proposed DeepVigor+ framework introduces a scalable and semi-analytical method for evaluating the fault resilience of CNNs. It offers a powerful alternative to traditional FI techniques. DeepVigor+ enables rapid and accurate reliability assessments of deep and complex CNNs at the early stages of design. This capability is particularly impactful for emerging AI and DL systems, where fast prototyping and early validation are critical for both research and deployment.

The analytical theories of DeepVigor+ open promising directions for extending reliability analysis beyond CNNs. Emerging DL models such as Transformers are increasingly gaining traction in various domains, including natural language processing, vision, and multimodal learning. Therefore, there is a growing need for efficient reliability assessment tools. The theory developed in DeepVigor+, particularly the optimal fault propagation analysis, can be generalized to these architectures with minimal overhead, primarily by integrating their unique operations into the framework. This adaptability makes DeepVigor+ a foundational tool for reliability analysis in the next generation of DL systems.

DeepVigor+ can play a significant role in quantifying the reliability of a DL system. DeepVigor+ provides the probabilistic metrics (i.e., VF) for failure due to faults at the DL model. Therefore, by combining platform-specific fault models and deployment constraints, DeepVigor+ could be integrated into design-time tools for co-optimizing efficiency and reliability. By developing cross-layer analysis, from algorithm to hardware implementation, it is possible to quantify how hardware architecture and mappings affect the reliability of the DL system. This comprehensive view facilitates the path toward certifying DL systems under emerging safety and reliability standards, such as those used in automotive or healthcare applications.

Beyond reliability analysis, DeepVigor+ can be exploited for designing fault-tolerant CNNs. Identifying the most vulnerable channels and layers by DeepVigor+ can lead to selective error detection and correction for the most vulnerable components in a DNN accelerator. While redundancy is quite costly for DNN accelerators, selective mechanisms can enable cost-effective fault tolerance.

Looking ahead, future research directions include extending DeepVigor+ to handle stochastic and temporal DL models, integrating it with real-world deployment environments (e.g., edge or neuromorphic hardware), and incorporating reliability metrics into Neural Architecture Search (NAS). These directions will further consolidate DeepVigor+ as a necessary component in building robust, safe, and certifiable AI systems.

%\vspace{-5mm}

\section{Conclusion} \label{sec:conclusion}

This paper addresses one of the major challenges in fault resilience analysis for CNNs. It introduces DeepVigor+, a semi-analytical, scalable alternative method to fault injection, quantifying deep CNNs' resilience accurately in a short time. DeepVigor+ is facilitated by a theoretical fault propagation modeling in CNNs, accompanied by stratified sampling, tackling the scalability problem for their resilience analysis. This open-source method enables fast resilience assessment, providing fine-grain evaluation and design space exploration for various fault-tolerant and cost-effective designs. 

The results in the paper indicate that DeepVigor+ derives vulnerability factors for layers and the entire model of CNNs with $14.9$ up to $26.9$ times fewer simulations than the best-known state-of-the-art statistical FI under similar less than $1\%$ error constraints. It is shown that CNN Model Vulnerability Factors can be obtained within minutes by analyzing their activations and weights. DeepVigor+ is also made public as an open-source tool for researchers and engineers to enable them to exploit it for CNNs' fault resilience assessment and enhancement. %\vspace{-10mm}

%% file: main.bbl
\begin{thebibliography}{10}
\providecommand{\url}[1]{#1}
\csname url@samestyle\endcsname
\providecommand{\newblock}{\relax}
\providecommand{\bibinfo}[2]{#2}
\providecommand{\BIBentrySTDinterwordspacing}{\spaceskip=0pt\relax}
\providecommand{\BIBentryALTinterwordstretchfactor}{4}
\providecommand{\BIBentryALTinterwordspacing}{\spaceskip=\fontdimen2\font plus
\BIBentryALTinterwordstretchfactor\fontdimen3\font minus \fontdimen4\font\relax}
\providecommand{\BIBforeignlanguage}[2]{{%
\expandafter\ifx\csname l@#1\endcsname\relax
\typeout{** WARNING: IEEEtran.bst: No hyphenation pattern has been}%
\typeout{** loaded for the language `#1'. Using the pattern for}%
\typeout{** the default language instead.}%
\else
\language=\csname l@#1\endcsname
\fi
#2}}
\providecommand{\BIBdecl}{\relax}
\BIBdecl

\bibitem{wang2022artificial}
Y.~Wang and S.~H. Chung, ``Artificial intelligence in safety-critical systems: a systematic review,'' \emph{Industrial Management \& Data Systems}, vol. 122, no.~2, pp. 442--470, 2022.

\bibitem{rech2024artificial}
P.~Rech, ``Artificial neural networks for space and safety-critical applications: Reliability issues and potential solutions,'' \emph{IEEE Transactions on Nuclear Science}, 2024.

\bibitem{athavale2020ai}
J.~Athavale, A.~Baldovin, R.~Graefe, M.~Paulitsch, and R.~Rosales, ``Ai and reliability trends in safety-critical autonomous systems on ground and air,'' in \emph{2020 50th Annual IEEE/IFIP International Conference on Dependable Systems and Networks Workshops (DSN-W)}.\hskip 1em plus 0.5em minus 0.4em\relax IEEE, 2020, pp. 74--77.

\bibitem{moghaddasi2023dependable}
I.~Moghaddasi, S.~Gorgin, and J.-A. Lee, ``Dependable dnn accelerator for safety-critical systems: A review on the aging perspective,'' \emph{IEEE Access}, 2023.

\bibitem{AIACT}
``Proposal for a regulation of the european parliament and of the council laying down harmonised rules on artificial intelligence (artificial intelligence act) and amending certain union legislative acts,'' \url{https://data.consilium.europa.eu/doc/document/ST-5662-2024-INIT/en/pdf}, 2024, [Online].

\bibitem{USACT}
``H.r.6216 - national artificial intelligence initiative act of 2020,'' \url{https://https://www.congress.gov/bill/116th-congress/house-bill/6216}, 2020, [Online].

\bibitem{rausand2014reliability}
M.~Rausand, \emph{Reliability of safety-critical systems: theory and applications}.\hskip 1em plus 0.5em minus 0.4em\relax John Wiley \& Sons, 2014.

\bibitem{ahmadilivani2023deepvigor}
M.~H. Ahmadilivani, M.~Taheri, J.~Raik, M.~Daneshtalab, and M.~Jenihhin, ``Deepvigor: Vulnerability value ranges and factors for dnns’ reliability assessment,'' in \emph{2023 IEEE European Test Symposium (ETS)}.\hskip 1em plus 0.5em minus 0.4em\relax IEEE, 2023, pp. 1--6.

\bibitem{bolchini2024resilience}
C.~Bolchini, L.~Cassano, and A.~Miele, ``Resilience of deep learning applications: A systematic literature review of analysis and hardening techniques,'' \emph{Computer Science Review}, vol.~54, p. 100682, 2024.

\bibitem{ahmadilivani2024systematic}
M.~H. Ahmadilivani, M.~Taheri, J.~Raik, M.~Daneshtalab, and M.~Jenihhin, ``A systematic literature review on hardware reliability assessment methods for deep neural networks,'' \emph{ACM Computing Surveys}, vol.~56, no.~6, pp. 1--39, 2024.

\bibitem{baumann2002impact}
R.~Baumann, ``The impact of technology scaling on soft error rate performance and limits to the efficacy of error correction,'' in \emph{Digest. International Electron Devices Meeting,}.\hskip 1em plus 0.5em minus 0.4em\relax IEEE, 2002, pp. 329--332.

\bibitem{henkel2013reliable}
J.~Henkel, L.~Bauer, N.~Dutt, P.~Gupta, S.~Nassif, M.~Shafique, M.~Tahoori, and N.~Wehn, ``Reliable on-chip systems in the nano-era: Lessons learnt and future trends,'' in \emph{Proceedings of the 50th Annual Design Automation Conference}, 2013, pp. 1--10.

\bibitem{safari2022survey}
S.~Safari, M.~Ansari, H.~Khdr, P.~Gohari-Nazari, S.~Yari-Karin, A.~Yeganeh-Khaksar, S.~Hessabi, A.~Ejlali, and J.~Henkel, ``A survey of fault-tolerance techniques for embedded systems from the perspective of power, energy, and thermal issues,'' \emph{IEEE Access}, vol.~10, pp. 12\,229--12\,251, 2022.

\bibitem{yatbaz2023introspection}
H.~Y. Yatbaz, M.~Dianati, and R.~Woodman, ``Introspection of dnn-based perception functions in automated driving systems: State-of-the-art and open research challenges,'' \emph{IEEE Transactions on Intelligent Transportation Systems}, 2023.

\bibitem{yuan2021tokens}
L.~Yuan, Y.~Chen, T.~Wang, W.~Yu, Y.~Shi, Z.-H. Jiang, F.~E. Tay, J.~Feng, and S.~Yan, ``Tokens-to-token vit: Training vision transformers from scratch on imagenet,'' in \emph{Proceedings of the IEEE/CVF international conference on computer vision}, 2021, pp. 558--567.

\bibitem{hussain2022design}
H.~Hussain, P.~Tamizharasan, and C.~Rahul, ``Design possibilities and challenges of dnn models: a review on the perspective of end devices,'' \emph{Artificial Intelligence Review}, pp. 1--59, 2022.

\bibitem{desislavov2023trends}
R.~Desislavov, F.~Mart{\'\i}nez-Plumed, and J.~Hern{\'a}ndez-Orallo, ``Trends in ai inference energy consumption: Beyond the performance-vs-parameter laws of deep learning,'' \emph{Sustainable Computing: Informatics and Systems}, vol.~38, p. 100857, 2023.

\bibitem{mohaidat2024survey}
T.~Mohaidat and K.~Khalil, ``A survey on neural network hardware accelerators,'' \emph{IEEE Transactions on Artificial Intelligence}, 2024.

\bibitem{ibrahim2020soft}
Y.~Ibrahim, H.~Wang, J.~Liu, J.~Wei, L.~Chen, P.~Rech, K.~Adam, and G.~Guo, ``Soft errors in dnn accelerators: A comprehensive review,'' \emph{Microelectronics Reliability}, vol. 115, p. 113969, 2020.

\bibitem{dos2023understanding}
F.~F. Dos~Santos, L.~Carro, and P.~Rech, ``Understanding and improving gpus' reliability combining beam experiments with fault simulation,'' in \emph{2023 IEEE International Test Conference (ITC)}.\hskip 1em plus 0.5em minus 0.4em\relax IEEE, 2023, pp. 176--185.

\bibitem{garrett2024soft}
T.~Garrett, S.~Roffe, and A.~George, ``Soft-error characterization and mitigation strategies for edge tensor processing units in space,'' \emph{IEEE Transactions on Aerospace and Electronic Systems}, 2024.

\bibitem{mittal2020survey}
S.~Mittal, ``A survey on modeling and improving reliability of dnn algorithms and accelerators,'' \emph{Journal of Systems Architecture}, vol. 104, p. 101689, 2020.

\bibitem{shafique2020robust}
M.~Shafique, M.~Naseer, T.~Theocharides, C.~Kyrkou, O.~Mutlu, L.~Orosa, and J.~Choi, ``Robust machine learning systems: Challenges, current trends, perspectives, and the road ahead,'' \emph{IEEE Design \& Test}, vol.~37, no.~2, pp. 30--57, 2020.

\bibitem{su2023testability}
F.~Su, C.~Liu, and H.-G. Stratigopoulos, ``Testability and dependability of ai hardware: Survey, trends, challenges, and perspectives,'' \emph{IEEE Design \& Test}, vol.~40, no.~2, pp. 8--58, 2023.

\bibitem{canziani2016analysis}
A.~Canziani, A.~Paszke, and E.~Culurciello, ``An analysis of deep neural network models for practical applications,'' \emph{arXiv preprint arXiv:1605.07678}, 2016.

\bibitem{chen2019binfi}
Z.~Chen, G.~Li, K.~Pattabiraman, and N.~DeBardeleben, ``Binfi: An efficient fault injector for safety-critical machine learning systems,'' in \emph{Proceedings of the International Conference for High Performance Computing, Networking, Storage and Analysis}, 2019, pp. 1--23.

\bibitem{ruospo2023assessing}
A.~Ruospo, G.~Gavarini, C.~De~Sio, J.~Guerrero, L.~Sterpone, M.~S. Reorda, E.~Sanchez, R.~Mariani, J.~Aribido, and J.~Athavale, ``Assessing convolutional neural networks reliability through statistical fault injections,'' in \emph{2023 Design, Automation \& Test in Europe Conference \& Exhibition (DATE)}.\hskip 1em plus 0.5em minus 0.4em\relax IEEE, 2023, pp. 1--6.

\bibitem{PytorchFI2020}
A.~{Mahmoud}, N.~{Aggarwal}, A.~{Nobbe}, J.~R.~S. {Vicarte}, S.~V. {Adve}, C.~W. {Fletcher}, I.~{Frosio}, and S.~K.~S. {Hari}, ``Pytorchfi: A runtime perturbation tool for dnns,'' in \emph{2020 50th Annual IEEE/IFIP International Conference on Dependable Systems and Networks Workshops (DSN-W)}, 2020, pp. 25--31.

\bibitem{weng2024fkeras}
O.~Weng, A.~Meza, Q.~Bock, B.~Hawks, J.~Campos, N.~Tran, J.~M. Duarte, and R.~Kastner, ``Fkeras: A sensitivity analysis tool for edge neural networks,'' \emph{Journal on Autonomous Transportation Systems}, 2024.

\bibitem{khoshavi2020fiji}
N.~Khoshavi, C.~Broyles, Y.~Bi, and A.~Roohi, ``Fiji-fin: A fault injection framework on quantized neural network inference accelerator,'' in \emph{2020 19th IEEE International Conference on Machine Learning and Applications (ICMLA)}.\hskip 1em plus 0.5em minus 0.4em\relax IEEE, 2020, pp. 1139--1144.

\bibitem{taheri2023appraiser}
M.~Taheri, M.~H. Ahmadilivani, M.~Jenihhin, M.~Daneshtalab, and J.~Raik, ``Appraiser: Dnn fault resilience analysis employing approximation errors,'' in \emph{2023 26th International Symposium on Design and Diagnostics of Electronic Circuits and Systems (DDECS)}.\hskip 1em plus 0.5em minus 0.4em\relax IEEE, 2023, pp. 124--127.

\bibitem{taheri2024saffira}
M.~Taheri, M.~Daneshtalab, J.~Raik, M.~Jenihhin, S.~Pappalardo, P.~Jimenez, B.~Deveautour, and A.~Bosio, ``Saffira: a framework for assessing the reliability of systolic-array-based dnn accelerators,'' in \emph{2024 27th International Symposium on Design \& Diagnostics of Electronic Circuits \& Systems (DDECS)}.\hskip 1em plus 0.5em minus 0.4em\relax IEEE, 2024, pp. 19--24.

\bibitem{mukherjee2003systematic}
S.~S. Mukherjee, C.~Weaver, J.~Emer, S.~K. Reinhardt, and T.~Austin, ``A systematic methodology to compute the architectural vulnerability factors for a high-performance microprocessor,'' in \emph{Proceedings. 36th Annual IEEE/ACM International Symposium on Microarchitecture, 2003. MICRO-36.}\hskip 1em plus 0.5em minus 0.4em\relax IEEE, 2003, pp. 29--40.

\bibitem{chen2020tensorfi}
Z.~Chen, N.~Narayanan, B.~Fang, G.~Li, K.~Pattabiraman, and N.~DeBardeleben, ``Tensorfi: A flexible fault injection framework for tensorflow applications,'' in \emph{2020 IEEE 31st International Symposium on Software Reliability Engineering (ISSRE)}.\hskip 1em plus 0.5em minus 0.4em\relax IEEE, 2020, pp. 426--435.

\bibitem{laskar2022tensorfi+}
S.~Laskar, M.~H. Rahman, and G.~Li, ``Tensorfi+: a scalable fault injection framework for modern deep learning neural networks,'' in \emph{2022 IEEE International Symposium on Software Reliability Engineering Workshops (ISSREW)}.\hskip 1em plus 0.5em minus 0.4em\relax IEEE, 2022, pp. 246--251.

\bibitem{leveugle2009statistical}
R.~Leveugle, A.~Calvez, P.~Maistri, and P.~Vanhauwaert, ``Statistical fault injection: Quantified error and confidence,'' in \emph{2009 Design, Automation \& Test in Europe Conference \& Exhibition}.\hskip 1em plus 0.5em minus 0.4em\relax IEEE, 2009, pp. 502--506.

\bibitem{schorn2018accurate}
C.~Schorn, A.~Guntoro, and G.~Ascheid, ``Accurate neuron resilience prediction for a flexible reliability management in neural network accelerators,'' in \emph{2018 Design, Automation \& Test in Europe Conference \& Exhibition (DATE)}.\hskip 1em plus 0.5em minus 0.4em\relax IEEE, 2018, pp. 979--984.

\bibitem{wang2024enhancing}
J.~Wang, J.~Zhu, X.~Fu, D.~Zang, K.~Li, and W.~Zhang, ``Enhancing neural network reliability: Insights from hardware/software collaboration with neuron vulnerability quantization,'' \emph{IEEE Transactions on Computers}, 2024.

\bibitem{amarnath2022soft}
C.~Amarnath, M.~Mejri, K.~Ma, and A.~Chatterjee, ``Soft error resilient deep learning systems using neuron gradient statistics,'' in \emph{2022 IEEE 28th International Symposium on On-Line Testing and Robust System Design (IOLTS)}.\hskip 1em plus 0.5em minus 0.4em\relax IEEE, 2022, pp. 1--7.

\bibitem{amarnath2023error}
------, ``Error resilience in deep neural networks using neuron gradient statistics,'' \emph{IEEE Transactions on Computer-Aided Design of Integrated Circuits and Systems}, 2023.

\bibitem{sabih2021fault}
M.~Sabih, F.~Hannig, and J.~Teich, ``Fault-tolerant low-precision dnns using explainable ai,'' in \emph{2021 51st Annual IEEE/IFIP International Conference on Dependable Systems and Networks Workshops (DSN-W)}.\hskip 1em plus 0.5em minus 0.4em\relax IEEE, 2021, pp. 166--174.

\bibitem{mahmoud2020hardnn}
A.~Mahmoud, S.~K.~S. Hari, C.~W. Fletcher, S.~V. Adve, C.~Sakr, N.~Shanbhag, P.~Molchanov, M.~B. Sullivan, T.~Tsai, and S.~W. Keckler, ``Hardnn: Feature map vulnerability evaluation in cnns,'' \emph{arXiv preprint arXiv:2002.09786}, 2020.

\bibitem{chatterjee2014impact}
I.~Chatterjee, B.~Narasimham, N.~Mahatme, B.~Bhuva, R.~Reed, R.~Schrimpf, J.~Wang, N.~Vedula, B.~Bartz, and C.~Monzel, ``Impact of technology scaling on sram soft error rates,'' \emph{IEEE Transactions on Nuclear Science}, vol.~61, no.~6, pp. 3512--3518, 2014.

\bibitem{sullivan2021characterizing}
M.~B. Sullivan, N.~Saxena, M.~O'Connor, D.~Lee, P.~Racunas, S.~Hukerikar, T.~Tsai, S.~K.~S. Hari, and S.~W. Keckler, ``Characterizing and mitigating soft errors in gpu dram,'' in \emph{MICRO-54: 54th Annual IEEE/ACM International Symposium on Microarchitecture}, 2021, pp. 641--653.

\bibitem{yan2020single}
Z.~Yan, Y.~Shi, W.~Liao, M.~Hashimoto, X.~Zhou, and C.~Zhuo, ``When single event upset meets deep neural networks: Observations, explorations, and remedies,'' in \emph{2020 25th Asia and South Pacific Design Automation Conference (ASP-DAC)}.\hskip 1em plus 0.5em minus 0.4em\relax IEEE, 2020, pp. 163--168.

\bibitem{bushnell2004essentials}
M.~Bushnell and V.~Agrawal, \emph{Essentials of electronic testing for digital, memory and mixed-signal VLSI circuits}.\hskip 1em plus 0.5em minus 0.4em\relax Springer Science \& Business Media, 2004, vol.~17.

\bibitem{elliott2013quantifying}
J.~Elliott, F.~Mueller, F.~Stoyanov, and C.~Webster, ``Quantifying the impact of single bit flips on floating point arithmetic,'' North Carolina State University. Dept. of Computer Science, Tech. Rep., 2013.

\bibitem{zhan2021improving}
J.~Zhan, R.~Sun, W.~Jiang, Y.~Jiang, X.~Yin, and C.~Zhuo, ``Improving fault tolerance for reliable dnn using boundary-aware activation,'' \emph{IEEE Transactions on Computer-Aided Design of Integrated Circuits and Systems}, vol.~41, no.~10, pp. 3414--3425, 2021.

\bibitem{huang2021rethinking}
Z.~Huang, W.~Shao, X.~Wang, L.~Lin, and P.~Luo, ``Rethinking the pruning criteria for convolutional neural network,'' \emph{Advances in Neural Information Processing Systems}, vol.~34, pp. 16\,305--16\,318, 2021.

\bibitem{singh1996stratified}
R.~Singh, N.~S. Mangat, R.~Singh, and N.~S. Mangat, ``Stratified sampling,'' \emph{Elements of survey sampling}, pp. 102--144, 1996.

\bibitem{ahmadilivani2023enhancing}
M.~H. Ahmadilivani, M.~Taheri, J.~Raik, M.~Daneshtalab, and M.~Jenihhin, ``Enhancing fault resilience of qnns by selective neuron splitting,'' in \emph{2023 IEEE 5th International Conference on Artificial Intelligence Circuits and Systems (AICAS)}.\hskip 1em plus 0.5em minus 0.4em\relax IEEE, 2023, pp. 1--5.

\end{thebibliography}
